\documentclass[review]{elsarticle}

\usepackage{lineno}
\usepackage[colorlinks = true,
            linkcolor = blue,
            urlcolor  = blue,
            citecolor = blue,
            anchorcolor = blue]{hyperref}

\usepackage{enumitem}
\usepackage{algorithm}
\usepackage{algorithmic}
\usepackage{amssymb}
\usepackage{subfig}
\usepackage{graphicx}
\usepackage{caption}
\usepackage{textcomp}
\usepackage{multirow}
\usepackage{multicol}
\usepackage{makecell}
\usepackage{longtable}
\usepackage{lipsum}
\usepackage{tabularx}
\usepackage{float}
\usepackage{url}
\usepackage{dblfloatfix}
\usepackage{mathtools}
\usepackage{cuted}
\usepackage{xcolor}
\newcolumntype{L}[1]{>{\raggedright\let\newline\\\arraybackslash\hspace{0pt}}m{#1}}
\newcolumntype{C}[1]{>{\centering\let\newline\\\arraybackslash\hspace{0pt}}m{#1}}
\newcolumntype{R}[1]{>{\raggedleft\let\newline\\\arraybackslash\hspace{0pt}}m{#1}}

\usepackage[left=2.5cm, right=2.5cm, top=2.5cm]{geometry}

\journal{Journal of ABC}

\begin{document}

\begin{frontmatter}

\title{An in-depth comparison of methods handling mixed-attribute data for general fuzzy min-max neural network}

\author[mymainaddress]{Thanh Tung Khuat \corref{mycorrespondingauthor}}
\cortext[mycorrespondingauthor]{Corresponding author}
\ead{thanhtung.khuat@student.uts.edu.au}
\address[mymainaddress]{Advanced Analytics Institute, Faculity of Engineering and Information Technology, University of Technology Sydney, \\NSW, Australia}

\author[mymainaddress]{Bogdan Gabrys}
\ead{bogdan.gabrys@uts.edu.au}

% \author[mymainaddress,mysecondaryaddress]{Elsevier Inc}

% \author[mysecondaryaddress]{Global Customer Service\corref{mycorrespondingauthor}}
% \cortext[mycorrespondingauthor]{Corresponding author}

% \address[mymainaddress]{1600 John F Kennedy Boulevard, Philadelphia}
% \address[mysecondaryaddress]{360 Park Avenue South, New York}

\begin{abstract}
A general fuzzy min-max (GFMM) neural network is one of the efficient neuro-fuzzy systems for classification problems. However, a disadvantage of most of the current learning algorithms for GFMM is that they can handle effectively numerical valued features only. Therefore, this paper provides some potential approaches to adapting GFMM learning algorithms for classification problems with mixed-type or only categorical features as they are very common in practical applications and often carry very useful information. We will compare and assess three main methods of handling datasets with mixed features, including the use of encoding methods, the combination of the GFMM model with other classifiers, and employing the specific learning algorithms for both types of features. The experimental results showed that the target and James-Stein are appropriate categorical encoding methods for learning algorithms of GFMM models, while the combination of GFMM neural networks and decision trees is a flexible way to enhance the classification performance of GFMM models on datasets with the mixed features. The learning algorithms with the mixed-type feature abilities are potential approaches to deal with mixed-attribute data in a natural way, but they need further improvement to achieve a better classification accuracy. Based on the analysis, we also identify the strong and weak points of different methods and propose potential research directions. 
\end{abstract}

\begin{keyword}
General fuzzy min-max neural network \sep data encoding methods \sep categorical features \sep mixed-type features \sep incremental learning algorithm \sep agglomerative learning algorithm
\end{keyword}

\end{frontmatter}

%\linenumbers

\section{Introduction}
Fuzzy neural networks, which are a combination of the neural network and fuzzy theory, have provided a powerful modelling paradigm to deal with many problems in pattern recognition and control systems \cite{Zengqi96, Fuller00}. They can effectively handle and process various types of uncertainties existing in the pattern recognition and classification problems, especially those with complex class boundaries \cite{Liu19}. The fuzzy min-max (FMM) neural network is a special kind of the neuro-fuzzy system developed by Simpson for classification \cite{Simpson92} and clustering \cite{Simpson93}. There have been a large number of algorithms improving the limitations of the original fuzzy min-max neural networks. Among them, a general fuzzy min-max (GFMM) neural network is a significantly enhanced version of the original model. It combines both classification and clustering in a single framework. Furthermore, the GFMM model can accept the input patterns in the forms of crisp points or intervals, labeled or unlabeled ones.

In general, the GFMM models use hyperboxes as the fundamental representation and modelling concept. Each $n$-dimensional hyperbox is defined by its minimum and maximum points. It is either labeled and therefore representing a part of a given class or unlabelled representing a cluster of unlabelled data points. The size of the hyperbox in each dimension is limited by a parameter called maximum hyperbox size ($0 \leq \theta \leq 1$). The GFMM learning algorithm's purpose is to generate new or adjust the existing hyperboxes to cover input patterns. The learning algorithms of GFMM neural networks are classified into two groups \cite{Gabrys04}, i.e., incremental (online) learning \cite{Gabrys00} and agglomerative (batch) learning algorithms \cite{Gabrys02a}.

Although fuzzy min-max neural networks have been applied to many real-world applications \cite{Khuat20sur}, one of the their main practical restrictions is that all input patterns are expected to be numerical and continuously valued features \cite{Castillo12}. However, in practice, data are often represented by categorical attributes (features) that can take values from a finite set of unordered nominal elements \cite{Ienco12}. In many real-world applications, we often encounter datasets with mixed categorical and continuous features; for example, those taken from medical, social, and biological sciences, retailers, banks, and insurance firms \cite{Rezvan2013}. One of the simple ways is the substitution of categorical data with numerical values and processing them as continuous data. However, this operation implicitly defines a distance metric for groups, which may not be appropriate for many applications. It is due to the fact that the categorical variables may not possess continuity, and no relevant correspondence is shown between the continuous representations replaced in this manner and the original categorical values \cite{Brouwer02}. Another popular method for handling the categorical features is to employ one-hot encoding. However, with a large number of values for each feature, one-hot encoding usually increases the number of dimensions of the training sets, which leads to high requirements for computational resources. Moreover, the one-hot encoding method considers different values in the same categorical attribute entirely independently from each other. Therefore, it neglects the useful relations between values, which may affect the classification performance of the model. For instance, different values of a categorical feature \textit{City} may show the similarity based on the geographic distance. In addition to a label encoding and one-hot encoding, there are many other encoding methods based on various statistical measures which can be used to convert (encode) categorical features into numerical ones. However, as the impact of different types of encoding on the performance of GFMM classifiers have not been comprehensively evaluated in the past and it is not clear just from their definitions, it is one of the motivations for our study.

Since the original GFMM learning algorithms work effectively on numerical data, as an alternative to encoding of the categorical features, we could handle the classification problems on data with mixed features by introducing a hybrid system. In particular, the GFMM classifier could be applied to the numerical features only, while the categorical features could be handled by other effective learning algorithms, such as decision trees. Such an approach would require suitable aggregation of the predictive results from such two different models. We will discuss and evaluate the performance of such a method in the subsequent sections.   

The third approach to handle the datasets with mixed features in the GFMM model is to construct the learning algorithms able to tackle both types of features simultaneously during the training process. Although a lot of improved versions have been proposed since the original fuzzy min-max neural network was introduced, as shown in a recent survey paper \cite{Khuat20sur}, there are only two studies focusing on adapting some of the existing learning algorithms for datasets with mixed categorical and numerical features. The first study was proposed in \cite{Castillo12}, in which the authors extended the original GFMM classifier to accommodate categorical features by introducing a new membership function, new learning procedures for categorical features, and a new structure for the GFMM neural network. Nevertheless, the effectiveness of the proposed method was only assessed on the imputation of missing values for data in opinion polls. As a result, the performance of this classifier for classification problems has not been verified. In this paper, we will assess the efficiency of this classifier for the classification on the datasets with mixed features as well as those with only categorical attributes. The second study on building a classification algorithm for fuzzy min-max models on both numerical and categorical data was presented in \cite{Shinde16}. However, their idea has been applied to the modified fuzzy min-max neural network. In this paper, we bring this idea to the GFMM neural network and within a consistent, extensive experimental framework evaluate its effectiveness in comparison to other methods of handling the data with mixed features.

This paper aims to perform an in-depth comparison of the effectiveness of three main approaches to tackling data with only categorical or mixed categorical and numerical features applied to the GFMM family of pattern classifiers and learning algorithms. Based on the experimental results, we will discuss the advantages and disadvantages of each method. Through these analyses, we identify the existing problems when using the GFMM neural networks to process data with mixed categorical and continuous features and inform the potential directions in this field of research. Our main contributions in this paper can be summarized as follows:

\begin{itemize}
    \item We perform a comprehensive comparison of three principal techniques to tackle classification problems on the data with mixed categorical and numerical features.
    \item The strong and weak points when combining the methods with the GFMM neural network are identified and discussed via extensive experiments on different datasets.
    \item We elaborate on the existing problems of the applicability of the GFMM neural network to classification the datasets with mixed features and propose the potential research directions.
\end{itemize}

The rest of this paper is organized as follows. Section \ref{prolbemsolution} presents the motivation of this study and proposes the main approaches to handling the classification issues on the data with mixed categorical and numerical features. Section \ref{background} summarizes the main points of the GFMM neural network and its learning algorithms as well as feature encoding methods. Experimental results and discussion are shown in Section \ref{experiment}. Existing problems and potential directions are mentioned in Section \ref{direction}. Section \ref{conclusion} concludes the findings of this paper.

\section{Problems and solutions} \label{prolbemsolution}
\subsection{Problem statement}
Let $\mathbf{X}^{(Tr)} = \{[X_1^l, X_1^u, A_1, c_1], \ldots, [X_N^l, X_N^u, A_N, c_N]\}$ be the $N$ training samples, in which $X_i^l = (x_{i1}^l, \ldots, x_{in}^l)$ and $X_i^u = (x_{i1}^u, \ldots, x_{in}^u)$ are the $n$ numerical features represented as the lower bound $X_i^l$ and upper bound $X_i^u$ for the $i^{th}$ training sample, $A_i = (a_1^i, \ldots, a_r^i)$ are the $r$ categorical values corresponding to the $r$ categorical features of the $i^{th}$ training sample, $a_k^i$ is a categorical value of the $k^{th}$ categorical feature $A_k$ for the $i^{th}$ training sample, $a_k^i \in \mbox{DOM}(A_k) = \{a_{1k}, a_{2k}, \ldots, a_{n_k k}\}$, where $\mbox{DOM}(A_k)$ is a domain of categorical values for the categorical feature $A_k$ and $n_k$ is the number of categorical values of the feature $A_k$. This paper aims to evaluate different methods to train an effective GFMM model from $\mathbf{X}^{(Tr)}$.

The original learning algorithms of the GFMM neural network work only on numerical features as the corresponding membership values of an input pattern in the respective hyperboxes, which form a critical component of the learning algorithm, can only be computed for such numerical features. As the data in many practical problems is composed of a mixture of both numerical and categorical features, to make the GFMM neural network deal effectively with such problems, we aim to identify and evaluate the most efficient methods of learning from such data. To achieve this goal, this paper will use three different classes of approaches to construct GFMM models from the datasets with mixed categorical and numerical features. The first class of methods is based on the use of categorical features encoding methods to transform categorical values into numerical values. We aim to identify the most appropriate encoding methods for each type of GFMM learning algorithm. Our goal is also to investigate potential issues concerning the impacts of the encoders on the performance of the learning algorithms. The second class of evaluated methods divides the training set with mixed features into two disjoint parts. The first part consists of the training samples with only numerical features, while the second part includes the training patterns with only categorical features. We use the first part to train the GFMM model, while the second part will be used to train another type of model performing effectively on the categorical data. The third class of methods adopts the algorithms which can learn from both numerical and categorical features to train the GFMM model. The effectiveness and drawbacks of these learning algorithms will also be thoroughly investigated.

\subsection{The use of encoding methods for categorical features}
\begin{figure}[!ht]
    \centering
    \includegraphics[width=0.9\textwidth]{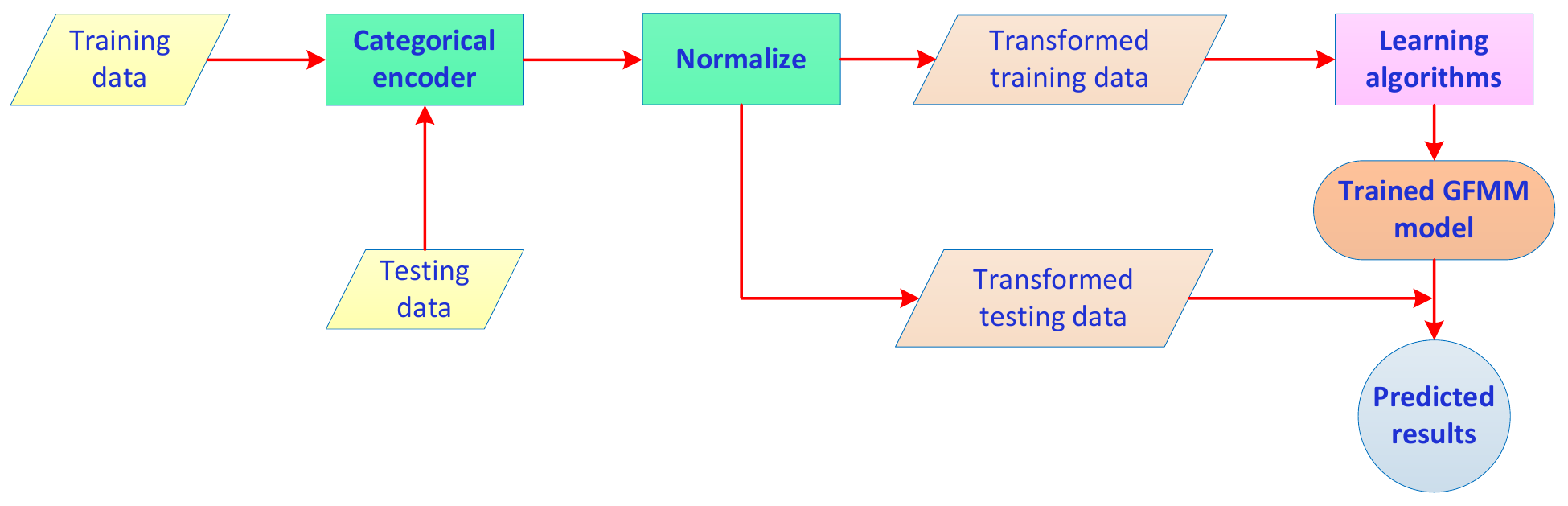}
    \caption{A diagram for the training and testing process of a GFMM model using categorical features encoding methods}
    \label{fig_encoding}
\end{figure}

In this type of approach, the training and testing sets are put through a single categorical encoder to transform the categorical values into numerical values. After that, encoded datasets are normalized into the range of [0, 1] as required the GFMM model and learning algorithms. Please note that the encoder and normalizer only use the information on the training set to transform both training and testing data. Next, the transformed training set is used to train a GFMM model. Finally, the performance of the classifier is evaluated on the transformed test data. The overall process of this approach is shown in Fig. \ref{fig_encoding}.

One of our goals in this study is to assess the impact of eight different categorical encoding methods, as shown in the next subsection \ref{encoding_methods}, on the performance of the GFMM classifier.

\subsection{Separate processing of numerical features and categorical features with the combination of GFMM model and another classifier}
This approach splits the original training dataset $\mathbf{X}^{Tr}$ into two parts. The first part contains the samples with numerical features only, and this part is used to train a GFMM model using the above learning algorithms. The second part consists of the patterns with only categorical features. We use this part to train a classifier working effectively on categorical features. In our experimental analyses, we used decision trees \cite{Quinlan93} which have been designed to naturally  deal with categorical features, though it has been frequently observed \cite{Cheng04} that  they can have problems with processing continuous features which need to be first discretized. Therefore, we propose to build a two-level hybrid model taking advantage of strong points of both the GFMM neural network and decision trees. From the training data, we can build a two-level hybrid classifier. This paper proposes two ways of combining the GFMM model and decision trees. The first approach uses only training data, while the second approach uses training and validation data.

\begin{figure}[!ht]
    \centering
    \includegraphics[width=0.8\textwidth]{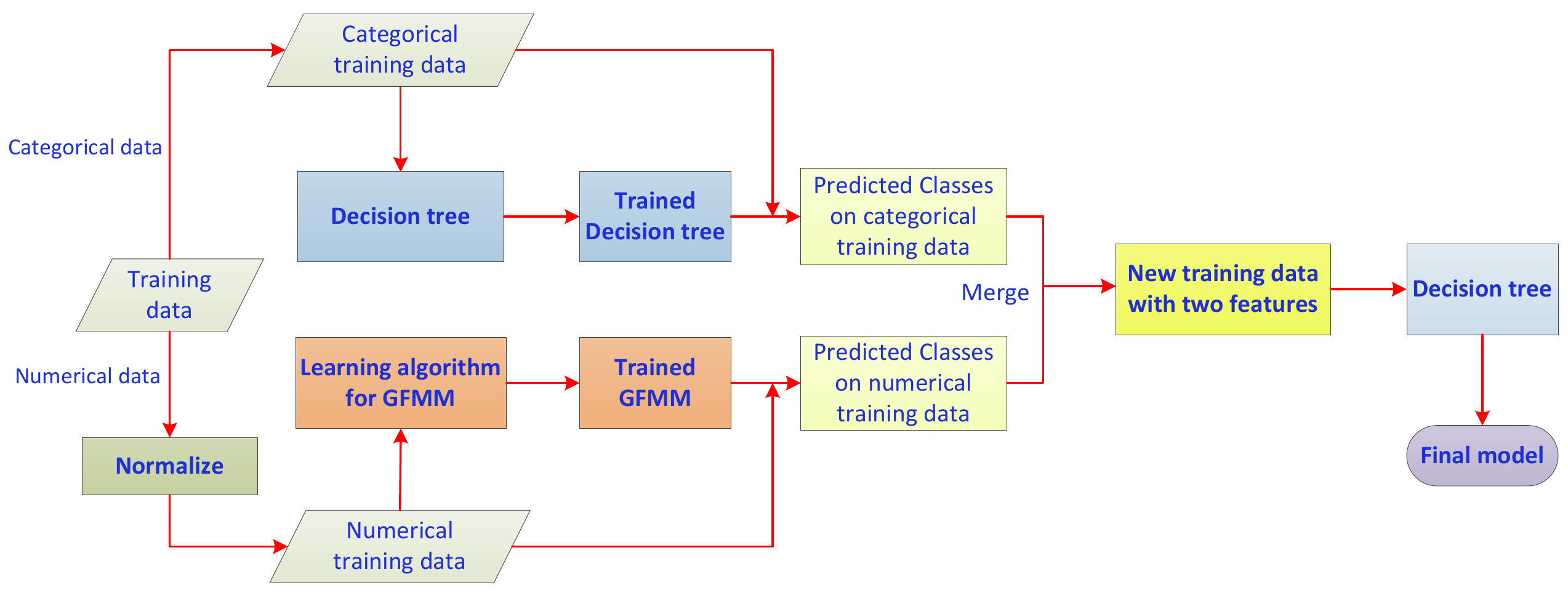}
    \caption{The first way of combination of GFMM model and decision tree using only training data}
    \label{fig_gfmm_dt_1}
\end{figure}

In the first approach, after training two models on the training set $\mathbf{X}^{Tr}$ separately, we fetch the same training data $\mathbf{X}^{Tr}$ to each model to obtain the predicted results. To prevent overfitting of the decision tree, we control the maximum depth of the decision tree. After receiving the predicted results from the GFMM neural network and decision tree on the training data, we append these two results to create a new training set containing two features. Finally, we use the new training data to train a new decision tree model. The overview of this method is presented in Fig. \ref{fig_gfmm_dt_1}.

\begin{figure}[!ht]
    \centering
    \includegraphics[width=0.8\textwidth]{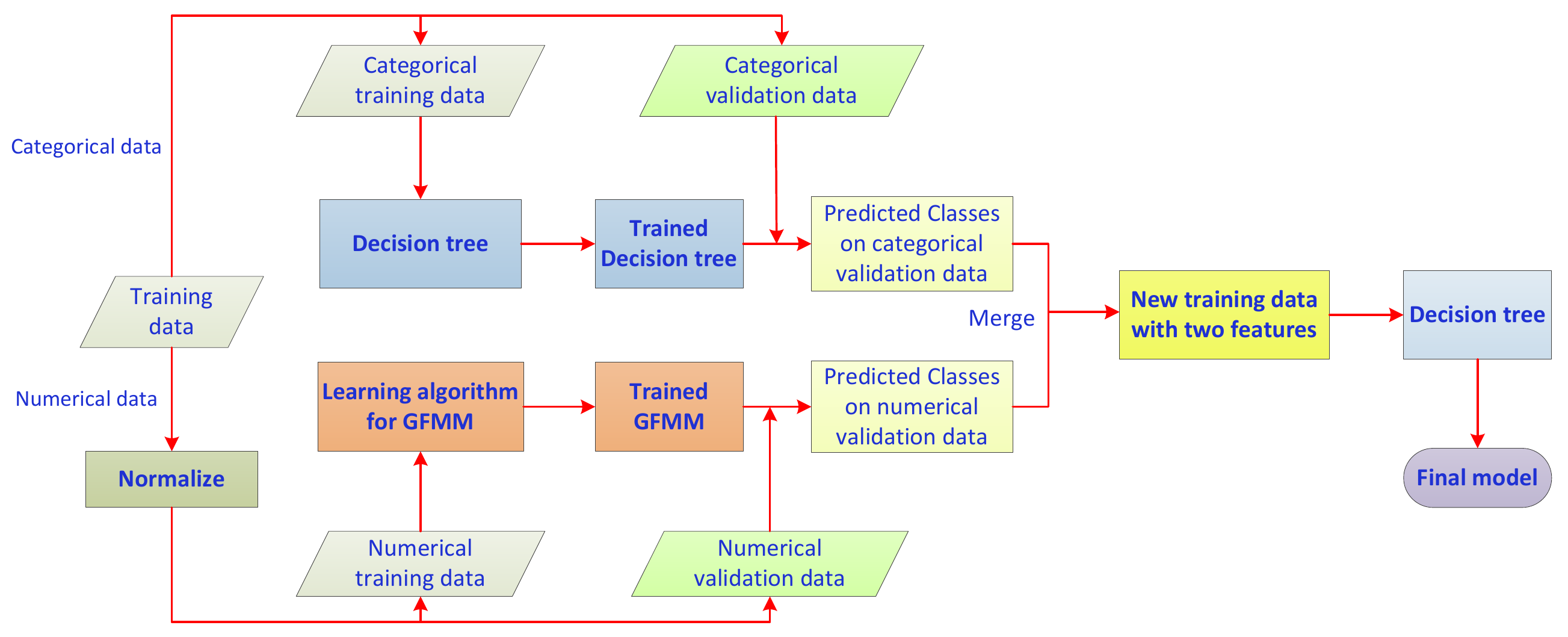}
    \caption{The second approach for combining the GFMM model and a decision tree using the training and validation data}
    \label{fig_gfmm_dt_2}
\end{figure}

The second method of combining the GFMM model and the decision tree uses the training data to train the first level of models and the validation data to train the second level of the hybrid classifier. The original training data is divided into two equal parts, i.e., training and validation parts. For the training part, the samples with only categorical features are employed to train the decision tree, while the patterns with only numerical attributes are put through the GFMM model. After obtaining the trained decision tree and GFMM model, we use the validation part to achieve the predicted results. The results are appended to form a new training set with two features, and these data are used to train another decision tree model. The main step of this model is shown in Fig. \ref{fig_gfmm_dt_2}.

\subsection{Employing specific learning mechanisms for categorical features}
The third approach for handling datasets with the mixed categorical and numerical features by the GFMM model is to use the learning algorithms that can deal effectively with both types of features. Up to now, there have been only two such learning algorithms proposed in the literature, as presented in subsection \ref{mix_alg}. Therefore, this paper compares these two learning algorithms with each other and with other ways of handling datasets with mixed categorical and numerical features. Please note that the numerical features need to be normalized to the range of [0, 1] before using the GFMM learning algorithms.

\section{Background} \label{background}
This section aims to provide the details of the structure of a GFMM neural network and its existing learning algorithms for only numerical features. Then, the extended versions of learning algorithms for mixed feature datasets are also described. Finally, we briefly present the encoding methods used to transform the categorical features into numerical features so that they can be employed in the original learning algorithms for only numerical, continuous features.

\subsection{General fuzzy min-max neural network}
The general fuzzy min-max neural network proposed in \cite{Gabrys00} includes three layers, i.e., input, hidden, and output layers. Let $n$ be the number of dimensions (features) in each input pattern; the input layer then comprises $2n$ nodes. The lower bounds of the input are stored in the first $n$ nodes, while the corresponding upper bounds for each feature are stored in the remaining $n$ nodes. Each input node is fully connected to $m$ hyperbox nodes in the middle layer. The connection weights of the lower bound nodes to the hyperboxes are kept in a matrix $\mathbf{V}$, while a matrix $\mathbf{W}$ stores the connection weights from the upper bound input nodes to hyperbox nodes. These values also represent the minimum and maximum points of hyperboxes, and they are tuned throughout the learning process. A membership function, also known as an activation function, for each hyperbox $B_i$ is defined by Eq. \eqref{membership}.
\begin{equation}
\small
    b_i(X) = \min\limits_{j = 1}^n(\min([1 - f(x_j^u - w_j, \gamma_j)], [1 - f(v_j - x_j^l, \gamma_j)]))
    \label{membership}
\end{equation}
where $ f(\lambda, \gamma) = \begin{cases} 
1, & \mbox{if } \lambda \gamma > 1 \\
\lambda \gamma, & \mbox{if } 0 \leq \lambda \gamma \leq 1 \\
0, & \mbox{if } \lambda \gamma < 0 \\
\end{cases}
$ is the ramp function, $ \gamma = [ \gamma_1,\ldots, \gamma_n ]$ is a sensitivity parameter controlling the slope of the membership function, and $X = [X^l, X^u]$ is the input sample given in a form of lower $X^l$ and upper $X^u$ bounds.

Each hyperbox $B_i$ is fully connected to each class node $c_k$ in the output layer by a binary-valued weight $u_{ik}$ as Eq. \eqref{eqhidout}. There are $p + 1$ class nodes corresponding to $p$ classes in the training set, where all unlabelled hyperboxes are linked to node $c_0$. The transfer function of each node $ c_k $ is the maximum membership degree of all hyperboxes connected to $c_k$ and is computed in Eq. \eqref{tranferfunc}.
\begin{equation}
\label{eqhidout}
    u_{ik} = \begin{cases}
        1, \quad \mbox{if hyperbox $B_i$ represents class $ c_k $} \\
        0, \quad \mbox{otherwise}
    \end{cases}
\end{equation}

\begin{equation}
    \label{tranferfunc}
    c_k = \max \limits_{i = 1}^m {b_i \cdot u_{ik}}
\end{equation}
where $m$ is the number of hyperboxes in the hidden layer. Although the GFMM neural network may be used for both types of labeled and unlabeled datasets, this paper is concerned with the fully labelled data and supervised classification problems only. As a result, the learning algorithms in the next sections are described for the classification.

\subsection{Learning algorithms for numerical, continuous features}
This part summarizes the main points of learning algorithms of the GFMM model for classification on the datasets with the numerical, continuous features. These include two incremental learning algorithms and an agglomerative learning algorithm. The accelerated versions of these algorithms using a novel hyperbox selection rule can be found in \cite{Khuat20acc}.

\subsubsection{Original online learning algorithm} \label{onln-gfmm}
The original incremental (online) learning algorithm for the GFMM classifier (Onln-GFMM) was proposed in \cite{Gabrys00}. The principal concept is to adjust or create new hyperboxes to cover the input patterns by a single scan through the training set. During the learning process, hyperboxes representing the same class are allowed to overlap, but the overlap between hyperboxes belonging to different classes is prohibited. In general, the learning algorithm includes three main steps, i.e., hyperbox expansion/generation, hyperbox overlap test, and hyperbox contraction.

Given training samples in the form of $X = [X^l, X^u, c_X]$, where $c_X$ is a class of $X$ while $X^l$ and $X^u$ are lower and upper bounds, the online learning algorithm first filters the hyperboxes with the same class as $c_X$ among all existing hyperboxes. Next, the membership values between $X$ and the selected hyperboxes are calculated and sorted in descending order. If there is at least one membership score with the value of one (i.e. the input pattern is fully contained within the core of one of the existing hyperboxes), the learning process continues with the next sample in the training set. In contrast, the algorithm will select in turn each hyperbox $B_i$ beginning from the hyperbox with the maximum membership degree to check the expansion condition. This operation ends when there is a hyperbox which can be expanded to include this input pattern. Otherwise, a new hyperbox with the same co-ordinates and class as $X$ is generated and added to the existing list of hyperboxes. The expansion condition for each hyperbox is shown in Eq. \eqref{expcondi}.
\begin{equation}
    \label{expcondi}
    \max(w_{ij}, x_j^u) - \min(v_{ij}, x_j^l) \leq \theta, \quad \forall{j \in [1, n]}
\end{equation}
where $\theta$ is the maximum hyperbox size in each dimension. If this condition is satisfied, hyperbox $B_i$ is expanded to cover $X$ employing Eqs. \eqref{expand} and \eqref{expand2}.
\begin{align}
    \label{expand}
    v_{ij}^{new} &= \min(v_{ij}^{old}, x_j^l) \\
     \label{expand2}
    w_{ij}^{new} &= \max(w_{ij}^{old}, x_j^u), \quad \forall{j \in [1, n]}
\end{align}

If the expansion procedure is performed, the extended hyperbox $B_i$ is tested for an overlap with hyperboxes $ B_k $ belonging to other classes. For each dimension $j$, four conditions are examined as follows (initially $\delta^{old} = 1$):

\begin{itemize}
    \item $v_{ij} \leq v_{kj} < w_{ij} \leq w_{kj}: \delta^{new} = \min(w_{ij} - v_{kj}, \delta^{old}) $
    \item $ v_{kj} \leq v_{ij} < w_{kj} \leq w_{ij}: \delta^{new} = \min(w_{kj} - v_{ij}, \delta^{old}) $
    \item $ v_{ij} < v_{kj} \leq w_{kj} < w_{ij}: \delta^{new} = \min(\min(w_{kj} - v_{ij}, w_{ij} - v_{kj}), \delta^{old}) $
    \item $ v_{kj} < v_{ij} \leq w_{ij} < w_{kj}: \delta^{new} = \min(\min(w_{ij} - v_{kj}, w_{kj} - v_{ij}), \delta^{old}) $
\end{itemize}

If $ \delta^{new} < \delta^{old} $, then we set $ \Delta = j$ and $ \delta^{old} = \delta^{new} $ to mark an overlap occurring on the $ \Delta{th} $ dimension, and the operation is iterated for the next dimension. Otherwise, no overlap exists between two considered hyperboxes ($\Delta = -1$), and the learning process continues with the next training sample. If $\Delta \ne -1$, the contraction procedure is adopted on the $\Delta{th}$ dimension to eliminate the overlap between two considered hyperboxes according to one of the following four cases:

Case 1: $ v_{i\Delta} \leq v_{k\Delta} < w_{i\Delta} \leq w_{k\Delta}: w_{i\Delta}^{new} = v_{k\Delta}^{new} = (w_{i\Delta}^{old} + v_{k\Delta}^{old}) / 2 $

Case 2: $ v_{k\Delta} \leq v_{i\Delta} < w_{k\Delta} \leq w_{i\Delta}:
        w_{k\Delta}^{new} = v_{j\Delta}^{new} = (w_{k\Delta}^{old} + v_{j\Delta}^{old}) / 2 $

Case 3: $ v_{i\Delta} < v_{k\Delta} \leq w_{k\Delta} < w_{i\Delta}: $
        \begin{align*}
        & v_{i\Delta}^{new} = w_{k\Delta}^{old}, \quad \mbox{if } w_{k\Delta} - v_{i\Delta} \leq w_{i\Delta} - v_{k\Delta} \\
        & w_{i\Delta}^{new} = v_{k\Delta}^{old}, \quad \mbox{if } w_{k\Delta} - v_{i\Delta} > w_{i\Delta} - v_{k\Delta}
    \end{align*}
        
Case 4: $ v_{k\Delta} < v_{i\Delta} \leq w_{i\Delta} < w_{k\Delta}: $
        \begin{align*}
        & w_{k\Delta}^{new} = v_{i\Delta}^{old}, \quad \mbox{if } w_{k\Delta} - v_{i\Delta} \leq w_{i\Delta} - v_{k\Delta} \\
        & v_{k\Delta}^{new} = w_{i\Delta}^{old}, \quad \mbox{if } w_{k\Delta} - v_{i\Delta} > w_{i\Delta} - v_{k\Delta}
    \end{align*}
    
In the classification phase, for each input sample the membership values to all existing hyperboxes in the trained model are computed. Next, the predicted class of the input pattern is the class of the hyperbox with the maximum membership value. However, in the original version of the GFMM model, if there are many winner hyperboxes with the same maximum membership score, we have to select the predictive class randomly among the classes of the winner hyperboxes. To cope with this issue, the experiments in this paper used a Manhattan distance from the input sample to the central points of the winner hyperboxes as shown in \cite{Upasani19}, and then the predicted class of the input pattern is the class of the winner hyperbox with the smallest distance value. If the input pattern is also a hyperbox itself, then we will compute the Manhattan distance from the central point of the input hyperbox to the central points of the winner hyperboxes.

\subsubsection{Agglomerative learning algorithm}
The original online learning algorithm depends on the order of the input patterns presentation because it creates or adjusts the hyperboxes in an incremental manner after receiving every single input sample. To overcome this drawback, Gabrys \cite{Gabrys02a} introduced an agglomerative (batch) learning algorithm using a full similarity matrix (AGGLO-SM). The AGGLO-SM algorithm begins with all training patterns representing hyperboxes and then merges the intra-class hyperboxes with the high similarity values and not generating the overlapping regions with the other hyperboxes belonging to different classes.

In the beginning, the AGGLO-SM algorithm builds the initial matrices $\mathbf{V}$ and $\mathbf{W}$ of minimum and maximum points by assigning to them the lower bounds \textbf{$X^l$} and upper bounds \textbf{$X^u$} of all input training patterns. After that, the algorithm carries out an iterated process of aggregating hyperboxes based on the computation and updating of the hyperbox similarity matrix. Three measures used for calculating the similarity value of each pair of hyperboxes can be found in \cite{Gabrys02a}. From the similarity matrix of intra-class hyperboxes, the algorithm merges hyperboxes sequentially by finding a pair of hyperboxes having the maximum similarity score larger than or equal to a minimum similarity threshold ($ \sigma $). After finding out such two hyperboxes (denoted by $B_i$ and $B_k$), the following constraints are checked before merging them:

\begin{enumerate}[label=(\alph*)]
    \item Maximum hyperbox size:
    $ \max(w_{ij}, w_{kj}) - \min(v_{ij}, v_{kj}) \leq \theta, \quad \forall j \in [1, n] $
    \item Non-Overlap. Newly merged hyperbox from $ B_i $ and $ B_k $ does not form overlapping regions with any existing hyperboxes representing other classes. The overlap test procedure is performed as in subsection \ref{onln-gfmm}. If there is any non-overlapping violation, another pair of hyperboxes is chosen and the steps are repeated as above.
\end{enumerate}

If all conditions are satisfied, the aggregation process of hyperboxes is performed as follows:
\begin{enumerate}[label=(\alph*)]
    \item Updating the minimum and maximum points of $ B_i $ to the coordinates of the merged hyperbox.
    \item Deleting $ B_k $ from the current set of hyperboxes and updating the similarity matrix.
\end{enumerate}

These learning operations are repeated until no pair of hyperbox candidates is found that can be merged.

It can be seen that the training process of the AGGLO-SM is time-consuming, especially in the case of large datasets, because we need to compute and sort the similarity matrix of pairs of hyperboxes. To decrease the training time, Gabrys \cite{Gabrys02a} proposed the second agglomerative algorithm (AGGLO-2), which reduces the computation of the full similarity matrix between all pairs of hyperboxes. The AGGLO-2 algorithm selects, in turn, each hyperbox in the current list of hyperboxes as an anchor hyperbox and performs the aggregation process among relevant hyperboxes. Assume that the first chosen hyperbox candidate is $ B_i $, the similarity scores between $ B_i $ and the remaining hyperboxes with the same class as $B_i$ are calculated. Hyperbox $ B_k $ with the highest similarity value is then selected as the second candidate for the merging process. The conditions and the merging process of $B_i$ and $B_k$ are the same as in the AGGLO-SM algorithm. If $B_i$ and $B_k$ do not satisfy the aggregation conditions, the hyperbox with the second-highest similarity score is chosen, and the above operations are iterated until the agglomeration process is performed or no pair of candidates for aggregation can be found.

In the classification phase, the predicted class of an input sample $X$ is the class of the hyperbox with the highest membership value for $X$. In the case of many hyperboxes representing $ K $ different classes with the same maximum membership value ($ b_{win}$), the predicted class of $X$ is the class $c_k$ with the highest value of $\mathbb{P}(c_k|X) $ defined as follows:
\begin{equation}
    \label{probcard}
    \mathbb{P}(c_k|X) = \cfrac{\sum_{j \in \mathcal{I}_{win}^k} n_j \cdot b_j}{\sum_{i \in \mathcal{I}_{win}} n_i \cdot b_i }
\end{equation}
where $k \in [1, K]$ and $ \mathcal{I}_{win} = \{ i, \mbox{if } b_i = b_{win} \}$ are a list of the indexes of all hyperbox with the same maximum membership value, $ I_{win}^k = \{ j, \mbox{if } class(B_j) = c_k \mbox{ and } b_j = b_{win} \}$ is a subset of $ I_{win} $ constructed from the indexes of the $k^{th}$ class, and $n_i$ is the number of patterns covered by a hyperbox $ B_i$.
 
 \subsubsection{An improved online learning algorithm}
 As demonstrated in \cite{Khuat2020iol}, the contraction process in the original online learning algorithm can lead to undesirable behaviour. Therefore, Khuat et al. \cite{Khuat2020iol} proposed an improved online learning algorithm (IOL-GFMM), which adopted the idea from the agglomerative learning algorithm introduced in \cite{Gabrys02a} where the overlap between hyperboxes from different classes is not allowed during the expansion of existing hyperboxes. Therefore, the learning in IOL-GFMM consists only the expansion/creation of hyperboxes and the overlap test.
 
 \paragraph{\textit{Hyperbox expansion and overlap testing step}} For each input training sample $ X = [X^l, X^u, c_X] $, all existing hyperboxes with the same class as $ c_X $ are selected and the membership values between $X$ and all selected hyperboxes are calculated. These membership scores are then sorted in the descending order. If the maximum membership degree is one, the process will continue with the next training sample. Otherwise, the IOL-GFMM algorithm selects, in turn, hyperbox candidates beginning with the hyperbox having the maximum membership degree, and then the expansion conditions are checked. If the hyperbox candidate meets all conditions, it will be expanded, and the training operation continues with the next training sample. If no candidate in the set of existing hyperboxes satisfies the expansion criteria, a new hyperbox is created to cover the input sample and appended to the current set of hyperboxes. Two expansion conditions are the maximum hyperbox size presented in Eq. \eqref{expcondi} and the non-overlap of the newly expanded hyperbox with any of the existing hyperboxes representing other classes.

The classification phase of the IOL-GFMM algorithm is performed the same as in the agglomerative learning algorithm.

\subsection{Existing GFMM learning algorithms for mixed features} \label{mix_alg}
\subsubsection{The first learning algorithm for mixed features}
The first study on extending the original learning algorithm \cite{Gabrys00} of the GFMM neural network for datasets with mixed categorical and numerical features can be found in \cite{Castillo12}. The authors used a similar representation of minimum ($V$) and maximum ($W$) points in the numerical features for categorical attributes, denoting them by $E$ and $F$. The $j^{th}$ categorical dimension of a hyperbox $B_i$ is determined by two categorical variables denoted by $e_{ij}$ and $f_{ij}$ with a full membership degree (a value of one). The authors proposed a distance measure between any two values of a categorical variable based on the occurrence probability of such categorical values with respect to the values of the class variable. In particular, the distance between two categorical values $a_{ij}$ and $a_{kj}$ of a categorical feature $A_j$ is defined as follows:

\begin{equation}
    \label{dist_cat}
    d(a_{ij}, a_{kj}) = \sqrt{\sum \limits_{c \in C} [\mathbb{P}(c|A_j = a_{ij}) - \mathbb{P}(c|A_j = a_{kj})]^2 }
\end{equation}
where $C$ is the set of all classes, $\mathbb{P}(c|A_j = a_{ij})$ is the conditional probability of class $c$ given $A_j = a_{ij}$. In practice, the conditional probability is unknown, and it is usually estimated from the training data as follows:
\begin{equation}
    \mathbb{P}(c|A_j = a_{ij}) = \cfrac{N_c(A_j = a_{ij})}{N(A_j = a_{ij})}
\end{equation}
where $N_c(A_j = a_{ij})$ is the number of training patterns having the feature value $a_{ij}$ for the feature $A_j$ and classified into class $c$, $N(A_j = a_{ij})$ is the number of training patterns with the categorical value $a_{ij}$ for the attribute $A_j$. Based on the distance between two categorical values, the membership function for a categorical value $a_{kj}$ on the $j^{th}$ categorical dimension $A_{j}$ of an input sample with respect to hyperbox $B_i$ is defined as follows:
\begin{equation}
    b_j(A_j=a_{kj}) = \min([1 - h(a_{kj}, e_{ij}), 1 - h(a_{kj}, f_{ij})])
\end{equation}
where $h(a_{kj}, e_{ij})$ is the normalized distance between two categorical values $a_{kj}$ and $e_{ij}$ and defined as follows:
\begin{equation}
    h(a_{kj}, e_{kj}) = \cfrac{d(a_{kj}, e_{ij})}{\max \limits_{k, i} d(a_{kj}, e_{ij})}
\end{equation}
It is noted that if $A_j = e_{ij}$ or $A_j = f_{ij}$, then $b_j = 1$. Based on those insights, the authors expanded the current membership function (Eq. \eqref{membership}) of a hyperbox $B_i$ with respect to the input sample including $n$ numerical features $[X^l, X^u] = ([x_1^l, x_1^u], \ldots, [x_n^l, x_n^u])$ and $r$ categorical features $A = (a_1, \ldots, a_r)$ as follows:
\begin{equation}
    b_i([X^l, X^u], A) = \min \{ \min\limits_{j = 1}^n[\min([1 - f(x_j^u - w_j, \gamma_j)], [1 - f(v_j - x_j^l, \gamma_j)])], \min\limits_{j = 1}^{r}[\min(1 - h_j(a_j, e_{ij}), 1 - h_j(a_j, f_{ij}))] \}
\end{equation}

The learning algorithm for the GFMM model including both categorical and numerical features, denoted Onln-GFMM-M1 in this paper, is modified from the original online learning algorithm. Note that the algorithm uses a default value $a_{0}$ to initialize the values for categorical bounds of the hyperbox ($h(a_{0}, a_{ij}) = h(a_{ij}, a_{0}) = 0$ for all categorical values $a_{ij}$). Assuming that the input pattern is in the form of $X = \{ [X^l, X^u], A, c_X \}$, where $X^l = (x_1^l,\ldots,x_n^l)$ and $X^u = (x_1^u,\ldots,x_n^u)$ are lower and upper bounds of $n$ numerical features, $A = (a_{1},\ldots,a_{r})$ contains the values of $r$ categorical features, the steps of the learning algorithm are shown as follows.

When the input training pattern $X$ comes to the network, the algorithm finds the hyperbox $B_i$ with the highest membership value and having the same class as $c_X$ to verify the expansion condition for both numerical and categorical features. If the maximum membership value is one, the algorithm continues with the next training sample. The expansion condition for the numerical features is the maximum hyperbox size ($\theta$) for each dimension shown in Eq. \eqref{expcondi}. If that condition is met, the algorithm will check the condition for each categorical dimension $j$ as follows:\\
\textit{Case 1:} If $e_{ij} = a_{0}$ and $f_{ij} = a_{0}$, then the condition for the categorical features is met without checking the remaining categorical dimension.\\
\textit{Case 2:} If $e_{ij} \ne a_{0}$ and $f_{ij} = a_{0}$, the following condition must be satisfied:
\begin{equation}
\label{eqmax_cat}
    h(e_{ij}, a_j) \leq \eta
\end{equation}
where $0 \leq \eta \leq 1$ is the maximum hyperbox size for the categorical features.\\
\textit{Case 3:} If $e_{ij} \ne a_{0}$ and $f_{ij} \ne a_{0}$, we first check whether the substitution either $e_{ij}$ or $f_{ij}$ with $a_j$ can increase the hyperbox size at the $j^{th}$ categorical dimension. If this constraint is met, then we will check the condition shown in Eq. \eqref{eqmax_cat} for the maximum hyperbox size after expanding the $j^{th}$ categorical dimension.

If the expansion conditions are met, the hyperbox $B_i$ will be expanded in both categorical and numerical features. The way of expanding the numerical features is shown in Eqs. \eqref{expand} and \eqref{expand2}, while each of the $j^{th}$ categorical feature is expanded as follows.\\
\textit{Case 1:} If $e_{ij} = a_{0}$ and $f_{ij} = a_{0}$, then $e_{ij} = a_j$\\
\textit{Case 2:} If $e_{ij} \ne a_{0}$ and $f_{ij} = a_{0}$, then $f_{ij} = a_j$\\
\textit{Case 3:} If $e_{ij} \ne a_{0}$ and $f_{ij} \ne a_{0}$ and $h(e_{ij}, a_j) > h(a_j, f_{ij})$, then $f_{ij} = a_j$\\
\textit{Case 4:} If $e_{ij} \ne a_{0}$ and $f_{ij} \ne a_{0}$ and $h(a_j, f_{ij}) > h(e_{ij}, a_j)$, then $e_{ij} = a_j$

If the expansion conditions are not met, the above process is repeated for the hyperbox with the next highest membership score. If no existing hyperbox can include or expand to cover the input pattern, a new hyperbox $B_i$ with the same class as $c_X$ is generated such that $V_i = X^l$, $W_i = X^u$, $E_i = A$, and $F_i = (a_{0}, \ldots)$.

If the selected hyperbox $B_i$ is expanded to cover the input pattern $X$, it has to be checked for overlap with hyperboxes representing other classes. The overlap test between numerical features is performed in the same way as in the original learning algorithm of the GFMM model. The overlap checking between two hyperboxes $B_i$ and $B_k$ on the $j^{th}$ categorical feature is verified as follows. If $e_{ij} = e_{kj}$ or $e_{ij} = f_{kj}$ or $f_{ij} = e_{kj}$ or $f_{ij} = f_{kj} \ne a_{0}$, then $B_i$ overlaps with $B_k$ on the $j^{th}$ categorical dimension. If all categorical and numerical features of two hyperboxes are overlapped with each other, a contraction process needs to be deployed to resolve the overlap on only one dimension causing the minimum change in the size of hyperboxes starting from the numerical features. We will replace a one categorical bound of a hyperbox by another value closer to the remaining categorical bound but not causing the overlap on that dimension with the remaining hyperbox. If this replacing operation is not possible for all categorical dimensions, we will deploy the contraction process for the numerical feature as steps in the original online learning algorithm shown in subsection \ref{onln-gfmm}. 

In the classification phase, the input pattern is classified into the class of the hyperbox with the maximum membership value. In case there are many winning hyperboxes with the same maximum membership value for an input pattern, we will compute the Manhattan distances from the input sample to the central point of the winning hyperboxes (using only numerical features). The predicted class of the input pattern is the class of the winning hyperbox with the smallest Manhattan distance. If the input pattern contains only categorical features, we will randomly select among classes of the winning hyperboxes to achieve the predicted class.

\subsubsection{The second learning algorithm for mixed features}
The second idea to build a learning algorithm for datasets with mixed categorical and numerical features was proposed in \cite{Shinde16}, but for the modified fuzzy min-max neural network. In this paper, we apply this idea to the general fuzzy min-max neural network denoted by Onln-GFMM-M2. Unlike the Onln-GFMM-M1 with each categorical feature of the hyperbox being specified by two categorical bounds, each categorical feature $j$ in the hyperbox of the Onln-GFMM-M2 algorithm is represented by a binary string $s_{ij}$, which is formed using the one-hot encoding technique for all values in the training set on the $j^{th}$ categorical dimension. Assuming that the training data contains $n$ numerical features and $r$ categorical features, each hyperbox $B_i$ in the Onln-GFMM-M2 is determined by a tuple of three vectors $V_i, W_i$, and $S_i$, where $V_i = \{v_{i1},\ldots,v_{in}\}$ and $W_i = \{w_{i1},\ldots,w_{in}\}$ are the minimum and maximum points for $n$ numerical features, $S_i = \{s_{i1},\ldots,s_{ir}\}$ is a vector containing $r$ binary strings representing $r$ categorical dimensions.

The membership function shown in Eq. \eqref{membership} is modified as Eq. \eqref{membership_oh_cat} for each input pattern $X = \{ [X^l, X^u], A, c_X \}$. Note that the categorical features of $X$ are coded using the one-hot encoding technique such that $A = \{a_1,\ldots,a_r\}$ with $a_j$ being a binary string for the $j^{th}$ categorical feature.
\begin{equation}
    \label{membership_oh_cat}
    b_i(X) = \frac{1}{2} \left[\min\limits_{j = 1}^n[\min(1 - f(x_j^u - w_j, \gamma_j), 1 - f(v_j - x_j^l, \gamma_j))] + \cfrac{1}{r} \sum \limits_{j=1}^r g(s_{ij}, a_j) \right]
\end{equation}
where the meaning of $f(\cdot, \cdot)$ and $\gamma$ are the same as in Eq. \eqref{membership} and $g$ is the membership function for the categorical features defined as follows:
\begin{equation}
    g(s_{ij}, a_j) = \begin{cases}
        1, \quad \mbox{if } bitand(s_{ij}, a_j) > 0 \\
        0, \quad \mbox{if } bitand(s_{ij}, a_j) = 0
    \end{cases}
\end{equation}
where $bitand(s_{ij}, a_j)$ is a bit-wise AND operation between two binary strings $s_{ij}$ and $a_j$.

The original learning algorithm of the GFMM model is changed for covering the categorical features. Similarly to the original algorithm, when an input pattern $X$ comes to the network, the algorithm computes the membership values between $X$ and all existing hyperboxes with the same class as $c_X$. If any membership value has the value of one, the learning algorithm continues with the next training sample. Otherwise, the algorithm sorts all membership values in descending order and selects a hyperbox $B_i$ with the highest membership value to check the expansion conditions. The expansion constraint for the continuous features is the same as in the Onln-GFMM algorithm shown in Eq. \eqref{expcondi}. The expansion condition for the categorical variables in $B_i$ is described as follows:
\begin{equation}
    \sum \limits_{j = 1}^r g(s_{ij}, a_j) \geq \beta 
\end{equation}
where $\beta$ is a user-defined parameter regularizing the minimum number of categorical values in the categorical features of the input pattern that match the values in the categorical dimensions of hyperbox $B_i$. If both expansion conditions are met, hyperbox $B_i$ is expanded to cover $X$. The expansion steps for the continuous attributes are shown in Eqs. \eqref{expand} and \eqref{expand2}. The categorical features of $B_i$ are updated as follows:
\begin{equation}
    s_{ij}^{new} = bitor(s_{ij}^{old}, a_j), \quad \forall{j \in [1, r]}
\end{equation}
where $bitor(s_{ij}^{old}, a_j)$ is a bit-wise OR operation between two binary strings $s_{ij}^{old}$ and $a_j$. If the expansion conditions for hyperbox $B_i$ are not satisfied, the hyperbox selection process finds the hyperbox with the next highest membership value and repeats the above steps until a hyperbox is expanded. If none of the existing hyperboxes can be expanded to accommodate $X$, a new hyperbox $B_i$ is created and added to the current set of hyperboxes such that $V_i = X^l$, $W_i = X^u$, and $S_i = A_i$.

If the selected hyperbox is expanded to cover the input pattern $X$, it will be checked for overlap with hyperboxes belonging to other classes and a contraction process will be performed if the overlap occurs. These steps are carried out for all continuous features of hyperboxes, and they are the same as in the original online learning algorithm.

The classification phase of this learning algorithm is the same as in the Onln-GFMM-M1 algorithm.

\subsection{Feature encoding methods}\label{encoding_methods}
This part presents several common feature encoding techniques used in this paper.
\subsubsection{Label encoding}
The label encoding method replaces categorical values in a form of string of characters by integer numbers so that learning algorithms can operate on these numbers easily. While label encoding is a simple way to transform categorical/symbolic data into numerical data, it imposes an order on the resulting data without considering explicitly the relationship between categorical values, which can lead to a reduction in the performance of learning models.
\subsubsection{One-hot encoding}
One-hot encoding is another simple method to deal with categorical variables. Unlike the label encoding, the results of applying one-hot encoding are binary rather than ordinal numbers. Formally, let $A_j$ be the $j^{th}$ categorical feature containing $k \geq 2$ categorical values such that $dom(A_j) = \{a_{lj}, 1 < l \leq k\}$, feature $A_j$ will be transformed into $k$ features, where each categorical value $a_{lj}^i$ at the $i^{th}$ sample of feature $A_j$ corresponds to a binary vector in a new feature space:
\begin{equation}
    a_{lj}^i = [\mathbf{1}(a_1^i = a_{lj}^i), \mathbf{1}(a_2^l = a_{lj}^i), \ldots, \mathbf{1}(a_k^i = a_{lj}^i)]
\end{equation}
where $a_k^i$ is the $k^{th}$ feature of the $i^{th}$ sample for the dimension $A_j$ in the new space, $\mathbf{1}(\cdot)$ is an indicator function. It is observed that, in this encoding method, each categorical value of a nominal variable is equidistant and orthogonal to each other. Hence, the one-hot encoding method is only appropriate in the case that categories are mutually exclusive \cite{Cohen03}.

\subsubsection{Sum Encoding}
Sum encoding, also known as effect encoding, is an encoding method for categorical values similar to the one-hot encoding, but it employs 0, 1, and -1 values for encoding features instead of only 0 and 1 in the one-hot encoding technique. For a categorical feature with $L$ categorical values, this method will generate $L - 1$ effect variables, denoted $e_1, \ldots, e_{L-1}$, to encode each categorical value. Among $L$ categorical values, the technique selects one value as a reference value. Assuming that the $L^{th}$ value is chosen, all effect variables encoding this $L^{th}$ value will be assigned the value of -1. For each $j^{th}$ categorical value among $L - 1$ remaining values, its $L - 1$ effect variables are coded as follows: $e_j = 1$ and $e_k = 0 \; (1 \leq k \leq L - 1, k \ne j)$.

\subsubsection{Helmert encoding}
If a categorical feature has $L$ categories, then the Helmert encoding method will create $L - 1$ new features to code the values of that categorical feature. Let $e_1, \ldots, e_{L-1}$ be $L-1$ encoded features for each categorical value. The first categorical value will be encoded by $L - 1$ values of -1. $L - 1$ remaining categorical values are encoded as follows. For the $j^{th}$ categorical value ($ j \geq 2$), we set $e_{j - 1} = j - 1; \; e_{i} = 0 \; (1 \leq i < j - 1); $ and $e_{k} = -1 \; (j \leq k \leq L - 1)$.

\subsubsection{Target encoding}
The target encoding method proposed in \cite{Micci-Barreca01} replaces each value of a categorical feature with a combination of the posterior probability of classes based on that categorical value and the prior probability of the classes computed from the training set. For a binary classification problem, each categorical value $a_k$ of a categorical feature $A_j$ is substituted by a scalar $s_k$ defined as follows:
\begin{equation}
\label{target_form}
    s_k = \lambda(N_k) \cfrac{N_{1k}}{N_k} + (1 - \lambda(N_k)) \cfrac{N_1}{N}
\end{equation}
where $N_{1k}$ is the number of samples having the categorical feature $A_j = a_k$ with a positive class $c = 1$, $N_k$ is the number of samples with $A_j = a_k$, $N_1$ is the number of training sample with class $c = 1$, and $N$ is the total number of training samples. $\lambda(N_k)$ is a monotonically increasing function on $N_k$ and gets the value in the range of [0, 1]. Weighting factor $\lambda(N_k)$ is formally defined as follows:
\begin{equation}
    \lambda(N_k) = \cfrac{1}{1 + e^{-\frac{N_k - m}{z}}}
\end{equation}
where $m$ determines the minimum number of samples for each category to obtain a reliable estimated result, and $z$ is a smoothing parameter controlling the transition between the posterior probability and the prior probability.

An extension of the target encoding from a two-class problem to a multi-class problem is straightforward, as shown in \cite{Micci-Barreca01} by identifying the value defined in Eq. \eqref{target_form} for each class. The target encoding allows us to retain useful information about categorical values in relation to classes but still keeping the same dimensionality as the original dataset.

\subsubsection{James–Stein encoding}
James–Stein encoding is an encoding method for categorical values based on the information of classes. This method is inspired by the James–Stein estimator \cite{James61}. The James-Stein encoder aims to shrink the average proportion of instances in each categorical group with respect to class variables towards the overall average proportion of the whole population. For the binary classification, each categorical value $a_k$ in the categorical variable $A_j$ is replaced by the proportion of samples with value $a_k$ belonging to positive class over the proportion of samples with the positive class. From the binary classification problem, the formulation can be easily extended for the multi-class classification problem. Formally, the James–Stein encoder replaces $a_k$ with the encoded value $\hat{a}_k$ defined as follows:
\begin{equation}
    \hat{a}_k = (1 - B) \cdot p_k + B \cdot p_{pos}
\end{equation}
where $p_{pos}$ is the overall proportion of the number of samples belonging to the positive class over the total number of training samples, $p_k$ is the proportion of the number of samples with $A_j = a_k$ and belonging to the positive class over the number of samples with $A_j = a_k$. $B$ is the weight of the population mean. There are many methods to find $B$, but one of them is to use the variance of the population and the variance of each group as follows:
\begin{equation}
    B = \cfrac{p_k \cdot (1 - p_k) / N_k}{p_k \cdot (1 - p_k) / N_k + p_{pos} \cdot (1 - p_{pos}) / N}
\end{equation}
where $N_k$ is the number of samples with $A_j = a_k$, and $N$ is the total number of training samples. We can see that when the variance of each group $a_k$ is much higher than the variance of the whole population, then $B \approx 1$, so the encoded value is biased to the proportion of the population. In contrast, when the variance of a group $a_k$ is much lower than that of the population variance, then $B \approx 0$, and so the encoded value is computed mainly from the proportion of that group.

\subsubsection{Leave-One-Out encoding}
Leave-One-Out (LOO) encoding method also uses the information of classes to form the encoded values. Assuming there are $N$ training samples, in which $a_j^i$ is the $i^{th}$ training sample of a categorical variable $A_j$ with the value $A_j = a_j^i$ and its class is $c_i$ ($c_i \in \mathbb{N}$), the LOO encoder will replace $a_j^i$ in the training set by an encoded real value $\hat{a}_j^i$ computed as follows:
\begin{equation}
    \hat{a}_j^i = \cfrac{\sum \limits_{t = 1}^N \mathbf{1}(a_j^t = a_j^i) \cdot c_t - c_i}{\sum \limits_{t = 1}^N \mathbf{1}(a_j^t = a_j^i) - 1}
\end{equation}
where $a_j^t$ is the value of the categorical feature $A_j$ at the $t^{th}$ training sample and $\mathbf{1}(a_j^t = a_j^i)$ is the indicator function, which returns 1 if $a_j^t = a_j^i$ and returns 0 if $a_j^t \ne a_j^i$.

As for testing samples, each categorical value $a_j^k$ of the categorical variable $A_j$ at the $k^{th}$ testing sample is substituted with the mean of the target of samples in the train set with $A_j = a_j^k$. It means that the categorical value $a_j^k$ in the testing set is replaced by $\hat{a}_j^k$ computed as follows:
\begin{equation}
\label{loo_test}
    \hat{a}_j^k = \cfrac{\sum \limits_{t = 1}^N \mathbf{1}(a_j^t = a_j^k) \cdot c_t}{\sum \limits_{t = 1}^N \mathbf{1}(a_j^t = a_j^k)}
\end{equation}

We can see that a categorical value $a_j$ can be encoded by different values in the training and testing sets. In addition, with the same categorical value $a_j$, there may be many different encoded values for different training samples.

\subsubsection{Catboost encoding}
Catboost encoding is an encoding method based on the information of the class variable, which was proposed in a recent study \cite{Prokhorenkova18}. It is similar to the LOO encoding approach, but the encoded value for a categorical value $a_j^i$ of the $i^{th}$ sample with class $c_i (c_i \in \mathbb{N})$ is computed based only on historical samples coming before that sample as follows:
\begin{equation}
\label{catboost_eq_train}
    \hat{a}_j^i = \cfrac{\sum \limits_{t = 1}^{i - 1} \mathbf{1}(a_j^t = a_j^i) \cdot c_t + z \cdot p}{\sum \limits_{t = 1}^{i - 1} \mathbf{1}(a_j^t = a_j^i) + z}
\end{equation}
where $a_j^t$ is the value of the categorical feature $A_j$ at the $t^{th}$ training sample and $\mathbf{1}(a_j^t = a_j^i)$ is the indicator function, which returns 1 if $a_j^t = a_j^i$ and returns 0 if $a_j^t \ne a_j^i$, $z > 0$ is an smoothing parameter. $p$ is a priori setting parameter, which is usually set to the mean of the target values in the training set.

Encoded values of the test data are computed in a similar manner as in the LOO encoding method using Eq. \eqref{catboost_test}:
\begin{equation}
\label{catboost_test}
    \hat{a}_j^k = \cfrac{\sum \limits_{t = 1}^N \mathbf{1}(a_j^t = a_j^k) \cdot c_t + z \cdot p}{\sum \limits_{t = 1}^N \mathbf{1}(a_j^t = a_j^k) + z}
\end{equation}
where $\hat{a}_j^k$ is the encoded value of $a_j^k$, $a_j^k$ is the categorical value of feature $A_j$ at the $k^{th}$ testing sample.

\section{Experimental results}\label{experiment}
\subsection{Datasets, parameter settings, and performance metrics}
To compare the performance of different methods of handling the datasets with mixed categorical and numerical features using the GFMM model, we performed the experiments on 14 datasets with mixed features or only categorical features taken from the UCI machine learning repository \footnote{https://archive.ics.uci.edu/ml/datasets.php}. A summary of these datasets including the numbers of samples, features, classes, and description of the domain of categorical features is presented in Table \ref{dataset}. 

\begin{table}[!ht]
\scriptsize{
\centering
\caption{Datasets were used for experiments.}\label{dataset}
\begin{tabular}{cL{2.2cm}C{1.4cm}C{1.2cm}C{1.7cm}C{1.8cm}L{5.5cm}}
\hline
ID & Dataset            & \# Samples & \# Classes & \# Numerical features & \# Categorical features & Description of  categorical features                                                                                               \\ \hline
1  & abalone            & 4177       & 28         & 7                     & 1                       & 1 feature: 3 values                                                                                                                \\ \hline
2  & australian         & 690        & 2          & 6                     & 8                       & 4 features: 2 values, 2 features: 3 values, 1 feature: 15   values, 1 feature: 8 values                                            \\ \hline
3  & cmc                & 1473       & 3          & 2                     & 7                       & 2 features: 4 values, 3 features: 2 values, 2 features: 4   values                                                                 \\ \hline
4  & dermatology        & 358        & 6          & 1                     & 33                      & 1 feature: 2 values, 1 feature: 3 values, 31 features: 4   values                                                                  \\ \hline
5  & flag               & 194        & 8          & 10                    & 18                      & 1 feature: 4 values, 1 feature: 6 values, 1 feature: 7 values,   1 feature: 10 values, 2 features: 8 values, 12 features: 2 values \\ \hline
6  & german             & 1000       & 2          & 7                     & 13                      & 4 features: 4 values, 3 features: 5 values, 1 feature: 10   values, 3 features: 3 values, 2 features: 2 values                     \\ \hline
7  & heart              & 270        & 2          & 7                     & 6                       & 3 features: 2 values, 1 feature: 4 values, 2 features: 3   values                                                                  \\ \hline
8  & japanese credit   & 653        & 2          & 6                     & 9                       & 4 features: 2 values, 3 features: 3 values, 1 feature: 14   values, 1 feature: 9 values                                            \\ \hline
9  & molecular biology & 3190       & 3          & 0                     & 60                      & All features: 4 values                                                                                                             \\ \hline
10 & nursery            & 12960      & 5          & 0                     & 8                       & 4 features: 3 values, 1 feature: 5 values, 2 features: 4   values, 1 feature: 2 values                                             \\ \hline
11 & post operative    & 87         & 3          & 1                     & 7                       & 5 features: 3 values, 2 features: 2 values                                                                                         \\ \hline
12 & tae                & 151        & 3          & 1                     & 4                       & 1 feature: 2 values, 1 feature: 25 values, 1 feature: 26   values, 1 feature: 2 values                                             \\ \hline
13 & tic tac toe      & 958        & 2          & 0                     & 9                       & All features: 3 values                                                                                                             \\ \hline
14 & zoo                & 101        & 7          & 1                     & 15                      & All features: 2 values                                                                                                             \\ \hline
\end{tabular}
}
\end{table}

For categorical features encoding methods, we used the library \textit{category-encoders}  \footnote{http://contrib.scikit-learn.org/category\_encoders/} (version 2.2.2) with default settings for the experiments. The experiments do not aim to compare the GFMM model to other classifiers. Instead, we only compare different methods to handle the datasets with mixed categorical and numerical features for the GFMM model. We will evaluate the impact of the introduced three classes of approaches on the classification performance of the GFMM model with different maximum hyperbox sizes ($\theta$). Therefore, we do not tune the maximum hyperbox size for different datasets but use the same maximum hyperbox size parameter for all experimental datasets. If we use a small value of the maximum hyperbox size, then the GFMM model is expected to have a high classification performance, but the complexity of the model is also high \cite{Khuat20comp}. In contrast, with large values of the maximum hyperbox size, the complexity of the GFMM model is low, but the classification performance is often not high. In this experiment, the different methods will be evaluated using three different thresholds of $\theta$, i.e., $\theta = 0.1, \theta = 0.7$, and $\theta = 1$. For several datasets, the encoded values in several dimensions contain only two values 0 and 1, so $\theta = 1$ will ensure that the hyperboxes can be expanded to cover both of these extreme values. For the agglomerative learning algorithm, we used the ``longest distance" \cite{Khuat20comp} as a similarity measure and set $\sigma = 0$ so that the performance of the learning algorithm depends only on the values of $\theta$. The sensitivity parameter $\gamma$ in the membership function impacts the decreasing speed of the membership degrees for the numerical features. If the high value of $\gamma$ is used, in some cases, there can be insufficient coverage of the space outside of the core hyperboxes and the maximum membership values for all hyperboxes are zero, which in turn usually leads to a misclassification. In the experiments, we employed $\gamma = 1$ as recommended in \cite{Abe01} and by doing so completely avoid the above described possibility. For the decision tree, to prevent overfitting, we set the value of the maximum depth of a tree to 10, as presented in \cite{Bertsimas17}. For the maximum hyperbox size of the categorical features in the Onln-GFMM-M1 algorithm, we used the values of $\eta = 0.1, \eta = 0.7$, and $\eta = 1$ similarly to $\theta$. As for the minimum number of categorical values matched between the input pattern and hyperbox candidate $\beta$ in the Onln-GFMM-M2 algorithm, we set $\beta$ to 25\%, 50\%, and 75\% of the number of categorical features for each dataset. 

The experiments in this paper were performed on multi-class imbalanced datasets, so we use the class balanced accuracy measure (CBA) to assess the performance of methods. As analysed in \cite{Mosley13} and \cite{Luque19}, CBA is an unbiased metric for assessing the performance of classifiers on the class imbalanced problems, and it can deal with the disadvantages of other metrics such as classification accuracy, balanced accuracy, Geometric mean, sensitivity, specificity \cite{Weng06} and Area Under the Curve (AUC) \cite{Valverde-Albacete10}.

The results of class balanced accuracy on each dataset in this section are the average of values on 10 times repeated 4-fold cross-validation.
\subsection{Evaluation of the impact of feature encoding methods on the performance of the GFMM neural network}
This experiment is to evaluate the impact of encoding methods on the performance of the learning algorithms of the GFMM neural network. Average class balanced accuracy results of the IOL-GFMM, Onln-GFMM, and AGGLO-2 algorithms using different encoding methods and only numerical features are shown in Tables \ref{table_encoded_iol}, \ref{table_encoded_onln}, and \ref{table_encoded_agglo2} in \ref{appendix_a} respectively. The best results in each row are highlighted in bold. In addition, the numbers of generated hyperboxes in the GFMM models using different encoding methods with the IOL-GFMM, Onln-GFMM, and AGGLO-2 algorithms are presented in Tables \ref{table_encoded_iol_hyperbox}, \ref{table_encoded_onln_hyperbox}, and \ref{table_encoded_agglo2_hyperbox} in \ref{appendix_a}.

In general, it is observed that the performance of the GFMM model using the mixed features with the encoding methods for the categorical features is usually higher than that using only numerical features for training. This fact indicates the important roles of categorical features with regard to the classification performance of the GFMM model. Therefore, categorical features should not be eliminated during the training process. The use of encoding methods to transform the categorical features into numerical features is a simple way to take advantage of useful information about categorical features for the training process. To facilitate the evaluation of the effectiveness of each encoding method on learning algorithms, we compute the average rank of encoding methods for three learning algorithms over 14 experimental datasets for each value of $\theta$. For each dataset and each threshold of $\theta$, the encoding method for a given learning algorithm leading to the highest average class balanced accuracy is assigned the first rank, the second-best is ranked two, and so on. Next, the average ranks of encoding methods for three learning algorithms at each threshold of $\theta$ are calculated, and these results are shown in Table \ref{avg_rank_encoding}. The rank of the best encoding method for each learning algorithm is shown in bold. 

\begin{table}[!ht]
\scriptsize{
\caption{Average rank of encoding methods over experimental datasets with different learning algorithms} \label{avg_rank_encoding}
\begin{tabular}{L{1.5cm}C{1.4cm}C{1.4cm}C{1.3cm}C{1.4cm}C{1.4cm}C{1.3cm}C{1.4cm}C{1.4cm}C{1.3cm}}
\hline
\multirow{2}{*}{\shortstack[l]{Encoding \\method}} & \multicolumn{3}{c}{$\theta = 0.1$} & \multicolumn{3}{c}{$\theta = 0.7$} & \multicolumn{3}{c}{$\theta = 1$} \\ \cline{2-10} 
 & IOL-GFMM   & Onln-GFMM   & AGGLO-2  & IOL-GFMM   & Onln-GFMM   & AGGLO-2  & IOL-GFMM  & Onln-GFMM  & AGGLO-2  \\ \hline
CatBoost                                                 & 12.4643          & 13.2857        & 13.1071          &    11.9286       &  14.1429         &   12.3571        & \textbf{9.7857} &     15.3571     &   11.2857           \\ \hline
One-hot                                                  &    19.3214        & 8.2857 &    19.6429       &     19.25      &   8.7143        &     18.8929      &    17.25       &    13.6786      &   16.9286        \\ \hline
LOO                                                      & 11.5 & 11.4643        & 10.8929          & 11.5 & 10.7857          & 12.3571          & 12.0714          & 10.8929         & 11.4643          \\ \hline
Label                                                    & 12.1786          & 11.6071        & 12.9643          & 12.8929          & 10.0357          & 12.8214          & 11.5714          & 10.8571         & 10.0357          \\ \hline
Target                                                   &    11.6071       &   \textbf{7.9643}      & \textbf{10.8214} & 10.8929          & 8.25 & \textbf{10.9643} &     10.9643      &   9.25       & 7.8929 \\ \hline
James-Stein                                              &    \textbf{11.3929}       &   8.8929      &   11.8214         &    \textbf{10.75}          &   \textbf{8.1071}        &   12.1071        &    10.8214       & \textbf{8.5357} &  \textbf{7.8214} \\ \hline
Helmert  &    18.6786       &   8.1429      &     18.6429      &       18.8214    &    8.2143       &      18.4643     &     16.3214        &    12.0714      &    17.25       \\ \hline
Sum                                                      & 13.25          & 9.0357        & 13.0357          & 13.6071          & 11.1786          & 12.9643          & 17.8929          & 11.4643         & 18.5357          \\ \hline
\end{tabular}
}
\end{table}

With the value of $\theta = 0.1$ and $\theta = 0.7$, the LOO, target, and James-Stein encoding methods are the ones leading to the best classification results for the learning algorithms without using the contraction process, i.e., IOL-GFMM and AGGLO-2. It is due to the fact that these encoding methods are based on a statistical measure generated from the relation between categorical values and classes. This information and its positive impact on the membership degree are maintained during the learning process when the hyperboxes are not disturbed by the contraction process. Target and James-Stein still contribute to the best classification performance for the Onln-GFMM algorithm regardless of using the contraction process. This confirms that the target and James-Stein encoding techniques are appropriate for the membership function in the GFMM neural network. Although the LOO encoding method also uses the information of classes to create the encoded values, it cannot help the Onln-GFMM algorithm obtain good predictive results as the shift in the encoded values between training and testing sets is influenced by the disturbance of hyperboxes due to the contraction step.

\begin{figure}
	\begin{subfloat}[Training set - CatBoost]{
			\includegraphics[width=0.48\textwidth, height=0.28\textheight]{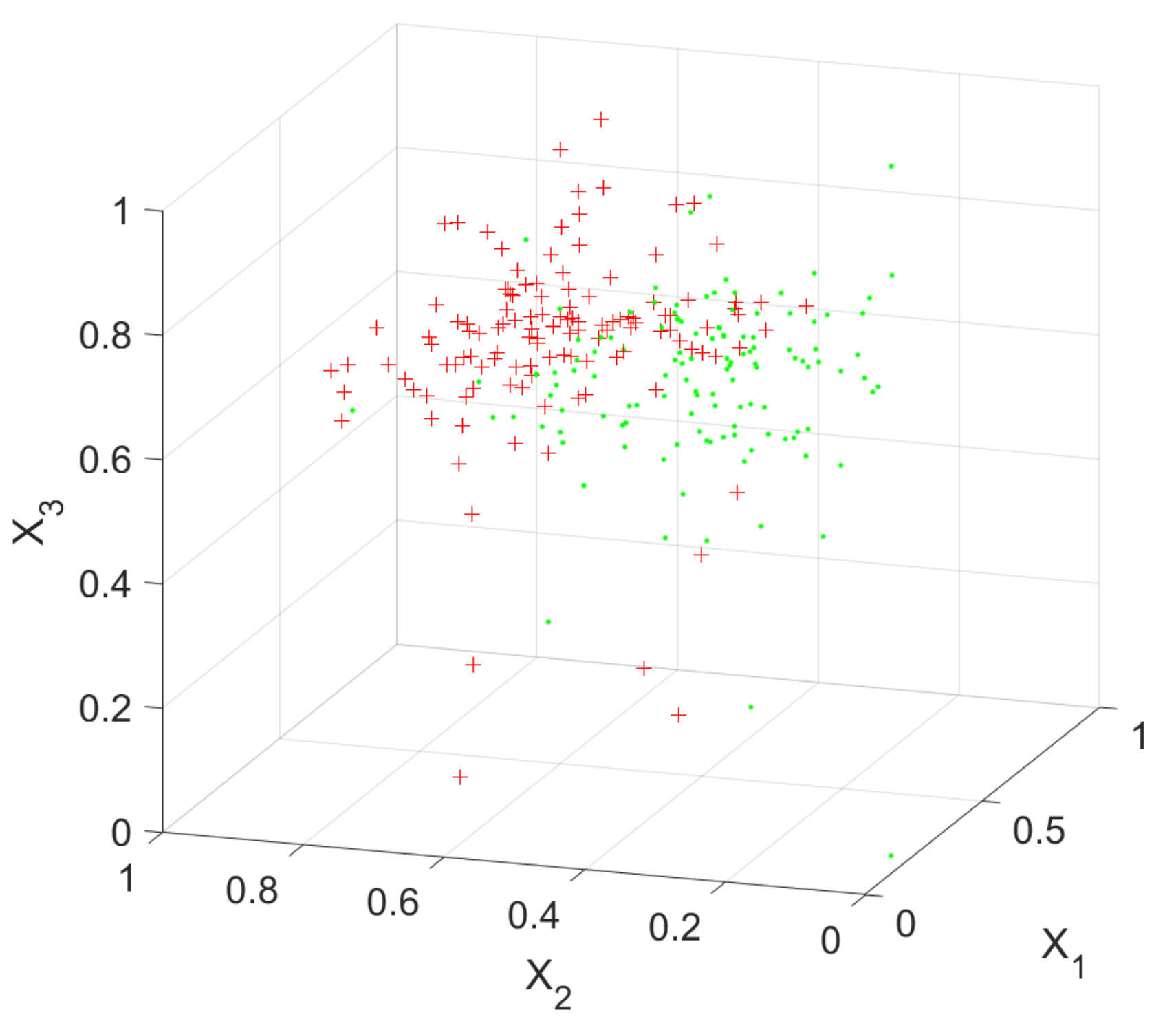}}
	\end{subfloat}
	\hfill%
	\begin{subfloat}[Testing set - CatBoost]{
			\includegraphics[width=0.48\textwidth, height=0.28\textheight]{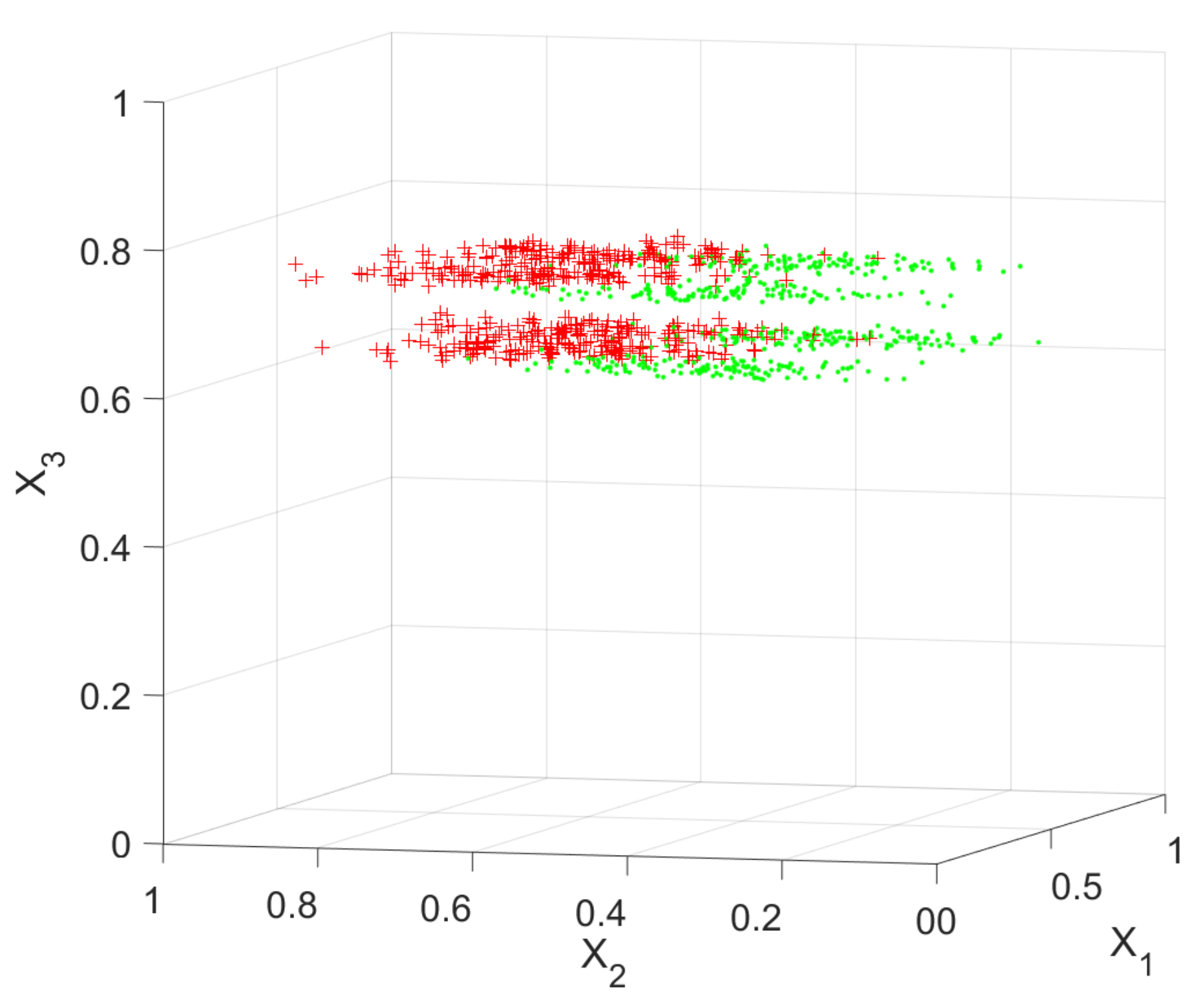}}
	\end{subfloat}
	\hfill%
	\begin{subfloat}[Training set - LOO]{
			\includegraphics[width=0.48\textwidth, height=0.28\textheight]{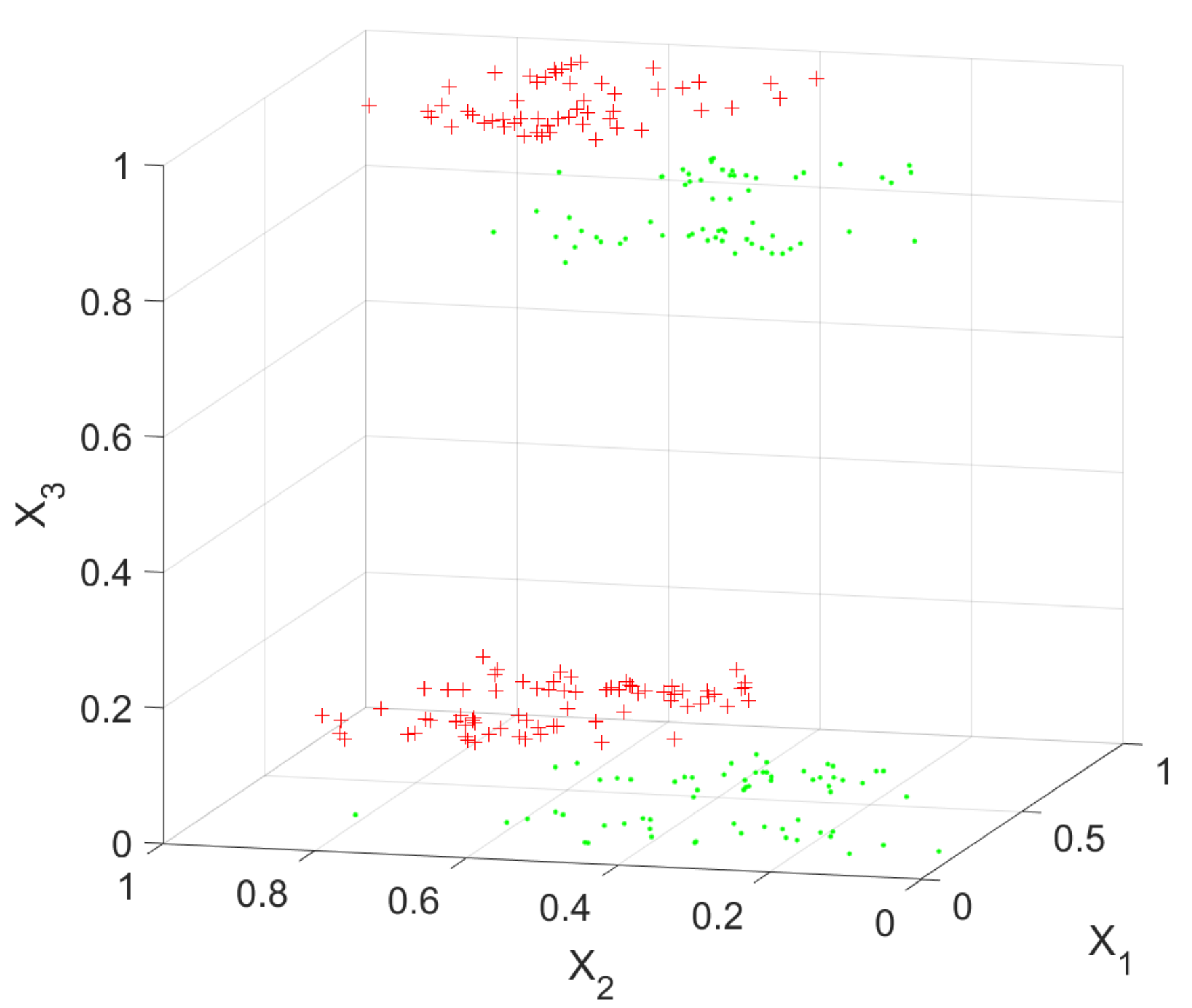}}
	\end{subfloat}
	\hfill%
	\begin{subfloat}[Testing set - LOO]{
			\includegraphics[width=0.48\textwidth, height=0.28\textheight]{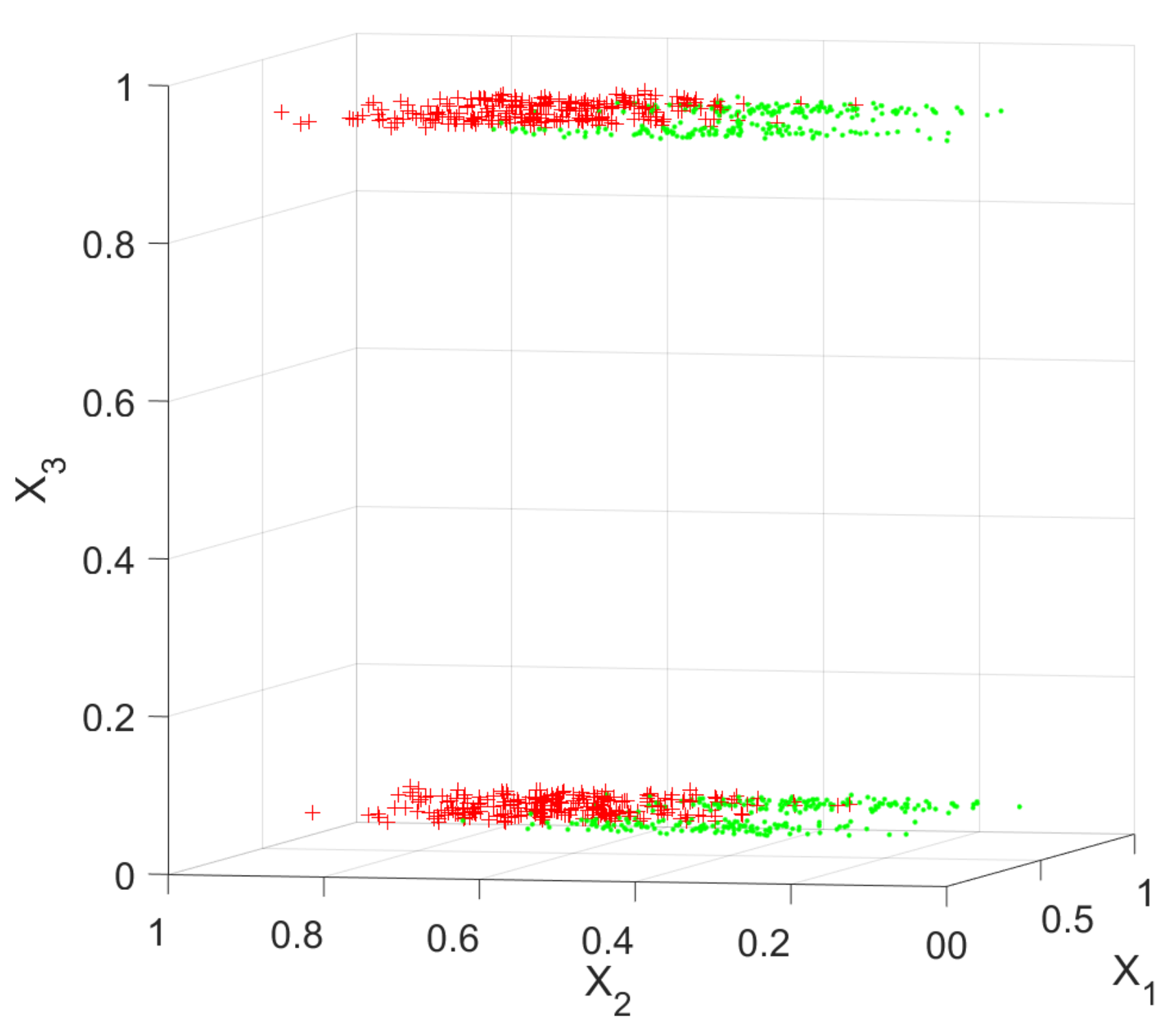}}
	\end{subfloat}
	\hfill%
	\begin{subfloat}[Training set - Target]{
			\includegraphics[width=0.48\textwidth, height=0.28\textheight]{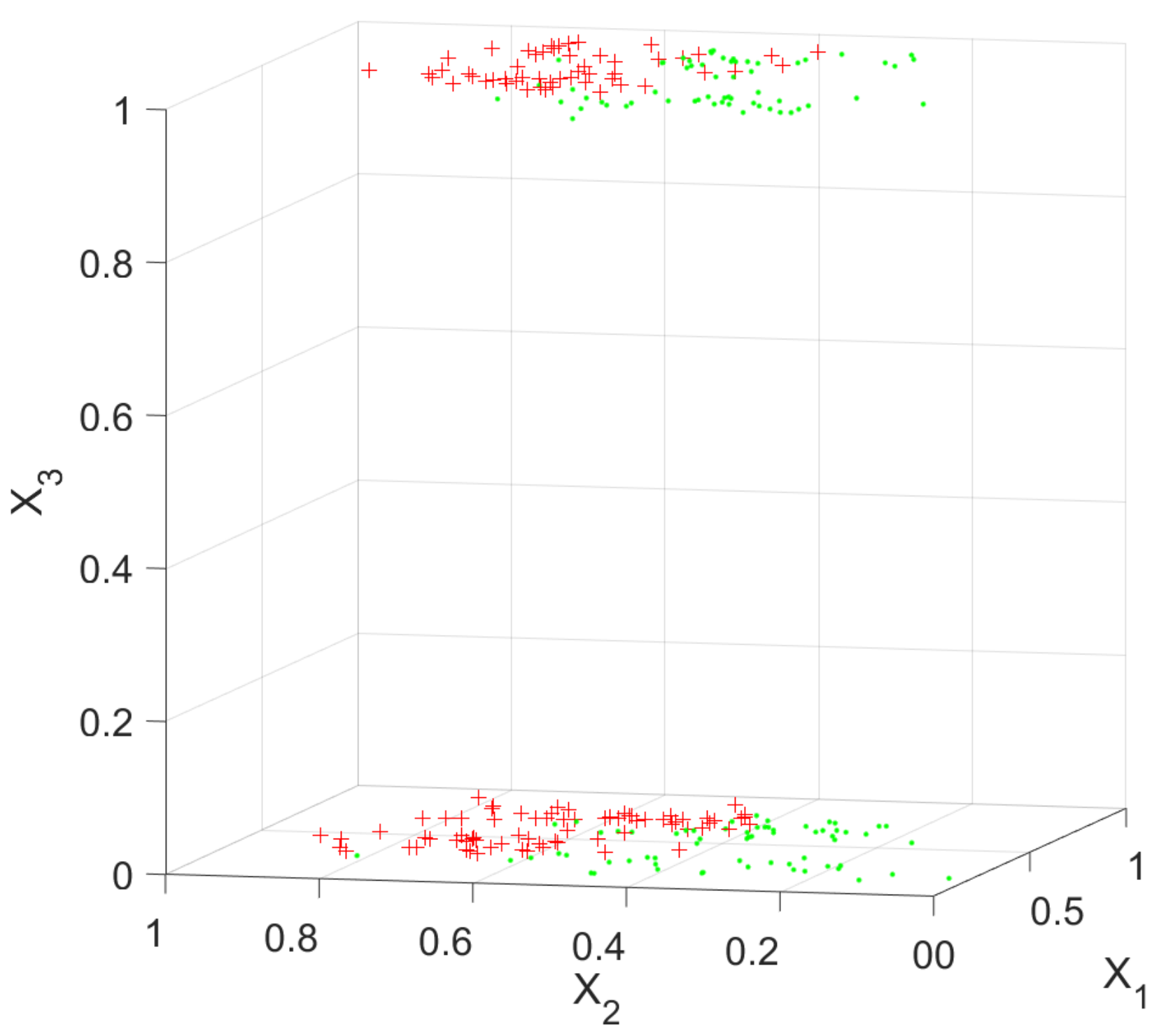}}
	\end{subfloat}
	\hfill%
	\begin{subfloat}[Testing set - Target]{
			\includegraphics[width=0.48\textwidth, height=0.28\textheight]{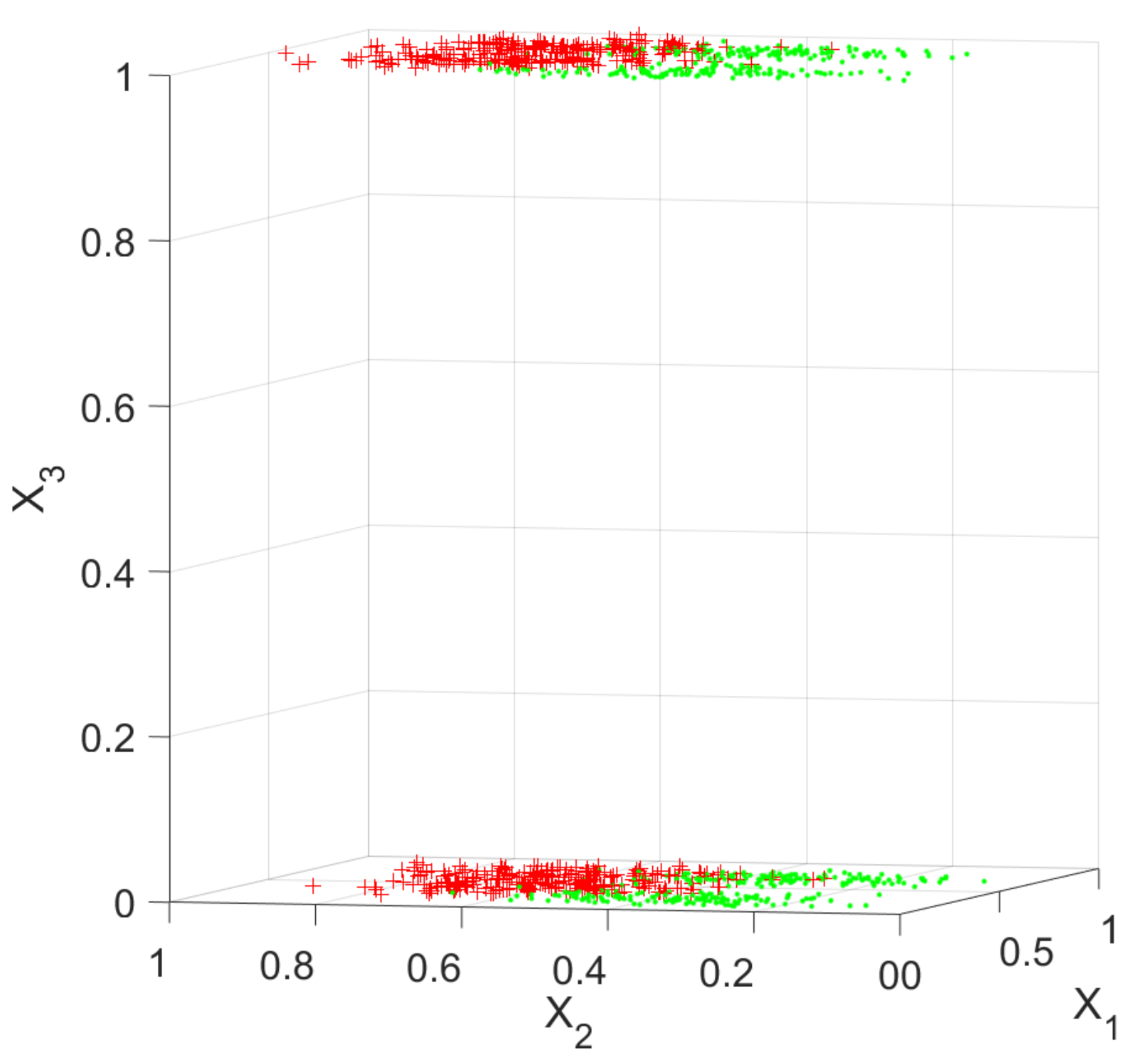}}
	\end{subfloat}
	\hfill
	\caption{The encoded values for the training and testing sets using CatBoost, LOO, and Target encoding methods with regard to the syntethic-1 dataset.}
	\label{demo_syn1}
\end{figure}

The CatBoost encoding method has not resulted in as good performance as the other statistical measure-based encoding methods for the IOL-GFMM algorithm with $\theta = 0.1$ and $\theta = 0.7$. The main reason for this result is that the shift in the encoded values between training and testing samples (as illustrated in Fig. \ref{demo_syn1}) makes the model with hyperboxes built on the training data impossible to perform well on the shifted testing samples. However, with $\theta = 1$, the performance of the IOL-GFMM algorithm using the CatBoost encoding method is the best among encoding methods. This is because the number of generated hyperboxes when using the CatBoost encoding approach is much higher than that using the other encoding methods, and so the resulting hyperboxes can capture the distribution of data better than the small number of hyperboxes created by the other encoding methods.

For the Onln-GFMM algorithm, the label encoding method usually results in poor classification performance because this encoding method imposes an artificial order among categorical values, and so the contraction process will cause the change in these artificial orders for the mix-max points of hyperboxes and in turn affecting the effectiveness of the membership function computed on the unseen input patterns with the unchanged artificial orders. As a result, the classification performance of the classifier is reduced. We can see that for the algorithms not using the contraction process such as IOL-GFMM and AGGLO-2, the classification accuracy of the model using the label encoding is close to the best predictive results. The CatBoost encoding method results in the worst classification performance of the GFMM model using the Onln-GFMM algorithm for all different considered values of $\theta$. This figure proves that the CatBoost encoding method has a negative impact on the classification ability of the Onln-GFMM algorithm, and it should not be used for the Onln-GFMM algorithm to train the GFMM neural network. There are two reasons for this poor classification accuracy. The first one is that the CatBoost encoding method depends on the order of the training data presentation, and thus it is severely affected by the contraction process in the Onln-GFMM algorithm. We can see that the CatBoost encoding method leads to high classification accuracy for the algorithms which do not use the contraction process, i.e., IOL-GFMM and AGGLO-2. The second reason is the shift of the encoded values between the training and testing data. As shown in Eqs \eqref{catboost_eq_train} and \eqref{catboost_test}, it is easily observed that with the same categorical value, the encoded value in the training data is different from the value in the testing data. Along with the contraction process, this offset causes incorrect predictive outcomes. This is also the reason to explain the poor predictive performance of the GFMM model trained using the Onln-GFMM algorithm with the LOO encoding method. From the experimental results, we strongly recommend that the encoding methods leading to the inconsistency of encoded values between training and testing samples such as LOO and CatBoost should not be deployed to encode the categorical values.

The one-hot or Helmert encoding method results in an excellent performance of the GFMM model using the Onln-GFMM algorithm with $\theta = 0.1$ or $\theta = 0.7$ because of the influence of Manhattan distance as a nearest neighbour measure in a highly complicated predictive model. We can see from Tables \ref{table_encoded_iol_hyperbox}, \ref{table_encoded_onln_hyperbox}, and \ref{table_encoded_agglo2_hyperbox} that the numbers of generated hyperboxes in the final model using the one-hot or Helmert encoding method are high for all learning algorithms compared to the models using other encoding approaches, and there are many hyperboxes containing only one or two training samples since the maximum hyperbox size condition is usually not satisfied with min-max points having values of 0 and 1. Hence, for each unseen sample, there are many winning hyperboxes because the values of 0 and 1 in the input pattern and min-max points of hyperboxes likely lead to zeros or ones (when the overlap areas cannot be detected by four current overlap test cases as shown in the next part) in the membership values. In the AGGLO-2 and IOL-GFMM algorithms, we use Eq. \eqref{probcard} to select the predicted class based on the number of samples included in each winning hyperbox. In this case, nonetheless, this formula can have the same probability value for different classes as the cardinality of each winning hypebox is small, and the selection process of the predicted class is performed randomly. As a result, the performance of the GFMM model using the IOL-GFMM or AGGLO-2 algorithm associated with the one-hot or Helmert encoding at $\theta = 0.1$ and $\theta = 0.7$ is worst in comparison to the model using the other encoding methods. Similar behaviour can also be observed in the case of the poor classification accuracy of the IOL-GFMM and AGGLO-2 algorithms with the use of the sum encoding method. In contrast, in the Onln-GFMM algorithm, we use the Manhattan distance to find the final winning hyperbox. With the min-max points of hyperboxes containing only values of 0 and 1 on categorical features, the hyperbox cannot be extended from 0 to 1 because of $\theta < 1$, so the maximum point and minimum point are the same. As a result, the central point on each categorical dimension also receives the same value as the min-max points for the categorical features. Therefore, the Manhattan distance from this central point to the input pattern including only the values of 0 and 1 on the categorical features can easily find the winning hyperbox closest to the input sample, and so the classification performance, in this case, is much higher than that using the probability formula as in the IOL-GFMM or AGGLO-2 algorithm. This fact is illustrated in many datasets such as \textit{australian}, \textit{dermatology}, \textit{flag}, \textit{german}, \textit{japanese credit}, \textit{nursery}, and \textit{zoo}. However, in the case of $\theta = 1$, each class is covered by only one hyperbox when using the Onln-GFMM algorithm with the one-hot encoding, and the contraction process is usually performed on the continuous features because it frequently causes less disturbance to the contracted hyperbox sizes than the case of doing it for the categorical features spanning the full range of its values. As a result, the sizes of a hyperbox in categorical features are usually kept unchanged. The large size of hyperboxes as well as many values of 0 and 1 in the encoded categorical features lead to high misclassification rates as the membership function can receive the value of zero easily when the encoded values on a given categorical feature mismatch between the unseen input pattern and the considered hyperbox.

To further reveal the impact of encoding methods on the learning algorithm and better explain why the One-hot and Helmert encoding methods can result in good classification results for Onln-GFMM algorithm but leading to bad predictive performance for AGGLO-2 and IOL-GFMM algorithms, we used synthetic three-dimensional datasets with two classes, in which two features are real numbers and the remaining feature is a categorical one. The data samples for each class with regard to numerical features have bimodal distribution as introduced in \cite{Ripley96}. The values of the categorical feature were randomly generated from a given set of values for samples in each class. We created two different datasets. The categorical feature of the first dataset called synthetic-1 contains only two values $\{One, Two\}$, while the categorical feature in the second dataset (synthetic-2) comprises 10 values in the set $\{One, Two, \ldots, Ten\}$. Each dataset contains 125 training samples and 500 testing samples for each class.

As mentioned above, CatBoost and LOO encoding methods result in the disturbance in the categorical values after performing the encoding step by generating many different encoded values for the same categorical value. We will illustrate this side-effect using the synthetic-1 dataset. The categorical feature in this dataset contains only two values. Therefore, other encoding methods except LOO and CatBoost will create only two encoded outcomes corresponding to two categorical values. However, LOO and CatBoost will generate many different encoded values for the same categorical value in the training set. Moreover, there is an offset among encoded values in training and testing sets in the categorical feature ($X_3$). Other remaining encoding methods such as Target do not results in such an undesired behaviour. These facets are demonstrated in Fig. \ref{demo_syn1} for the synthetic-1 dataset.

\begin{figure}
    \centering
    \includegraphics[width=0.6\textwidth]{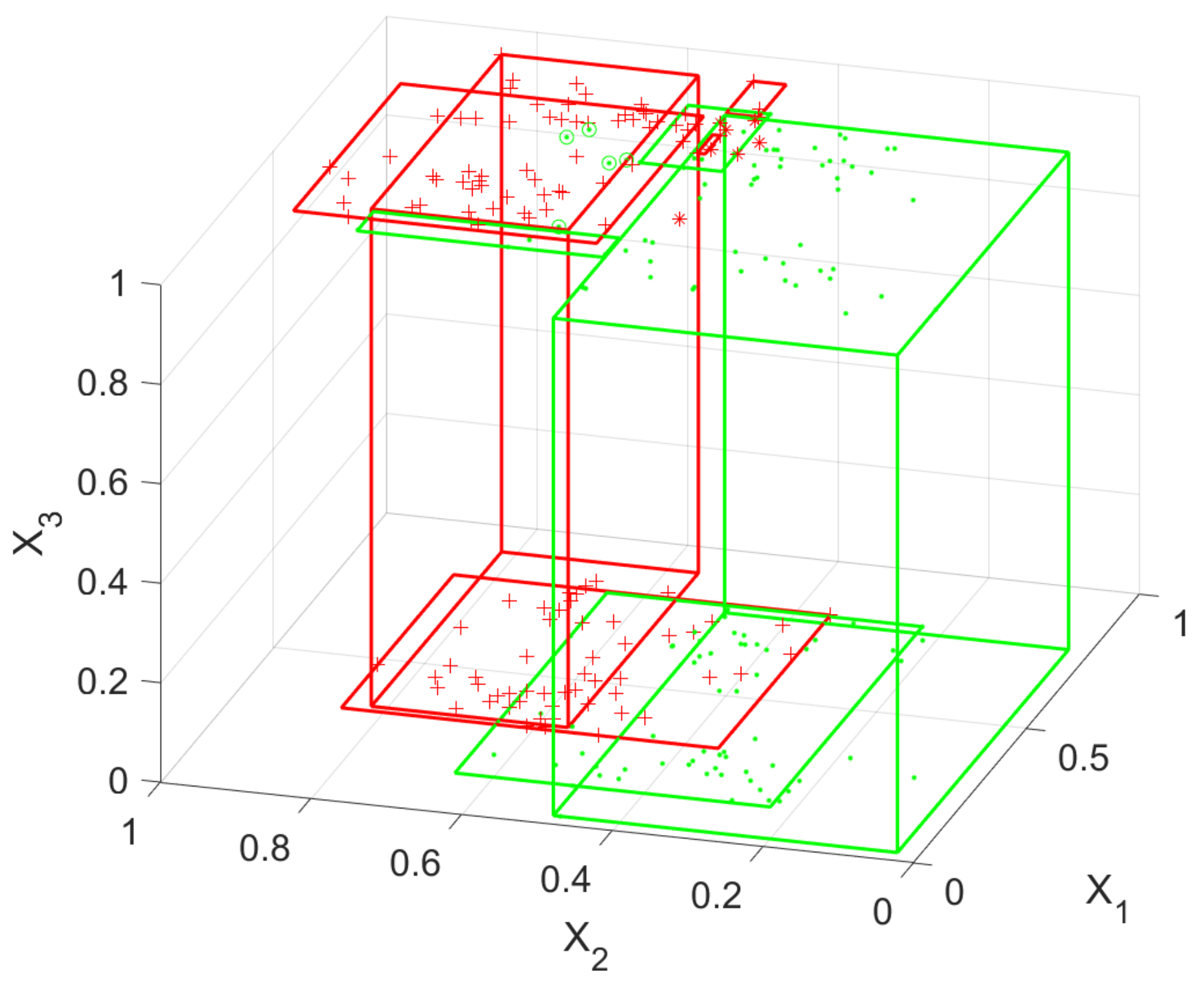}
    \caption{The generated hyperboxes of the IOL-GFMM algorithm using Helmert encoding method for synthetic-1 dataset ($\theta = 1$)}
    \label{demo_overlap}
\end{figure}

The classification accuracy, numbers of hyperboxes (No boxes), numbers of samples classified by the secondary criteria (the probability function for IOL-GFMM and AGGLO-2, Manhattan distance for Onln-GFMM) (denoted by No 2 criterion), numbers of samples classified correctly by using the secondary criterion (No 2 criterion correct) are shown in Tables \ref{table_syn_theta01},  \ref{table_syn_theta07}, and \ref{table_syn_theta1} in \ref{appendix_a} for all encoding methods and three learning algorithms with different values of $\theta$. From the obtained results, it can be easily observed that for large values of $\theta$, the number of patterns for which the secondary criterion beside membership function to find the appropriate class must be used is high for all learning algorithms using One-hot and Helmert encoding techniques. For the dataset with the categorical feature containing only two categorical values such as synthetic-1, the learning algorithms using target, label, James-Stein, and sum encoding methods also need to use the secondary criteria. The reason for this result stems from the drawback of the current overlap test cases. The four current overlap test cases cannot detect the overlapping regions between two hyperboxes $B_i$ and $B_k$ representing different classes in the case that $v_{ij} = v_{kj} = w_{ij} = w_{kj}$. For example, two hyperboxes with min-max points represented by $V_i = [0.3, 0.5], W_i = [0.4, 0.6], V_k = [0.35, 0.55], W_k = [0.45, 0.7]$ overlap in two-dimensional space. However, if their third categorical feature values are the same, e.g., $V_i = [0.3, 0.5, 0], W_i = [0.4, 0.6, 0], V_k = [0.35, 0.55, 0], W_k = [0.45, 0.7, 0]$, then these two hyperboxes are not considered as overlapping. This aspect is illustrated in Fig. \ref{demo_overlap} with the hyperboxes generated by the IOL-GFMM algorithm using Helmert encoding method for synthetic-1 dataset ($\theta = 1$). We can observe that many red and green hyperboxes overlap with each other when their $X_3$ values are the same.

The encoding methods such as one-hot or Helmert create many categorical dimensions with the same value (0 or 1) between two hyperboxes. In the case that a categorical feature contains only two categorical values, after normalising, the encoded data of many other encoding methods (except for LOO and CatBoost) shows only two values of 0 and 1 in the categorical dimension as well. Therefore, many hyperboxes with different classes can show the same value in the categorical dimensions, but these hyperboxes are not considered as overlapping in these categorical dimensions because the four existing test cases cannot detect this overlap case. As a result, IOL-GFMM and AGGLO-2 algorithms continue to expand or merge the existing hyperboxes which leads to the case of large overlapping areas when using the high value of $\theta$. Hence, many samples located in these overlapping areas show the same membership values, and we have to deploy an additional measure to classify the input patterns. This fact also occurs with the Onln-GFMM for $\theta < 1$. This shortcoming rarely happens for continuous features but it is a regular side-effect when encoding the categorical values. When the number of unique values in categorical features increases, this drawback usually decreases if we use encoding methods other than one-hot, Helmert, and sum. To reduce the negative impact of this drawback on the classification performance of learning algorithms, we can add more cases for the overlap test procedure or use the small values of $\theta$ or hyper-parameter tuning during the learning process.

From the experimental results, we can also see that the additional measure deployed for the case of many winning hyperboxes using Manhattan distance is more effective than the use of the probability function. Let us take the results of Onln-GFMM and IOL-GFMM algorithms for the synthetic-1 dataset with $\theta = 0.7$ (Table \ref{table_syn_theta07}) as an example. The numbers of generated hyperboxes and samples using the additional measure are the same for both learning algorithms. However, the number of samples classified correctly using Manhattan distance is much higher than that using the probability formula. As a result, the classification performance of the Onln-GFMM algorithm is higher than that using AGGLO-2 or IOL-GFMM with the one-hot, Helmert, and sum encoding methods. These experimental results on the synthetic datasets contribute to making the explanation mentioned in the case of real datasets more clearly.

To better comprehend the effectiveness of encoding methods on the learning algorithms of the GFMM neural network, we will use a statistical significance testing for the obtained results on the considered datasets. Our null hypothesis is:

\textit{H0: Given the GFMM learning algorithms using the same threshold of $\theta$, there is no difference in the classification performance of the GFMM models using different encoding methods}

To reject this hypothesis, we adopt a ``multiple testing" technique by the Friedman rank-sum test \cite{Eisinga17} and a posthoc test procedure as recommended in \cite{Demsar06}. Let $r_i^j$ be the rank of the $j^{th}$ classifier among $M$ classifiers on the $i^{th}$ dataset of $N$ datasets. In this experiment, $N = 14$ and $M = 24$. We calculate the Friedman statistic distribution based on average ranks $R_j$ as follows:
\begin{equation}
    \chi_F^2 = \cfrac{12 N}{M \cdot (M + 1)} \left[ \sum_{j=1}^M R^2_j - \cfrac{M \cdot (M + 1)^2}{4}\right]
\end{equation}

Based on this metric, Iman and Davenport \cite{Iman80} proposed a F-distribution with $M - 1$ and $(M - 1) \cdot (N - 1)$ degrees of freedom as follows:
\begin{equation}
    F_F = \cfrac{(N - 1) \cdot \chi_F^2}{N \cdot (M - 1) - \chi_F^2}
\end{equation}

The null hypothesis is rejected at the significance level $\alpha$ if $F_F$ is below a critical value
of $F(M - 1, (M-1)\cdot(N - 1), \alpha)$. If the null hypothesis is rejected, we need to use a posthoc test procedure to find the difference between pairs of methods. This paper will use the Nemenyi test and  Critical Difference (CD) diagram \cite{Demsar06} for the posthoc test.

With 14 datasets and eight encoding methods applied to three learning algorithms, $F_F$ in this experiment is distributed according to the F-distribution with $24 - 1 = 23$ and $(24 - 1) \cdot (14 - 1) = 299$ degrees of freedom. The critical value of $F(23, 299, 0.05)$ for the significance level $\alpha = 0.05$ is 1.5655.

\begin{figure}[!ht]
    \centering
    \includegraphics[width=0.85\textwidth]{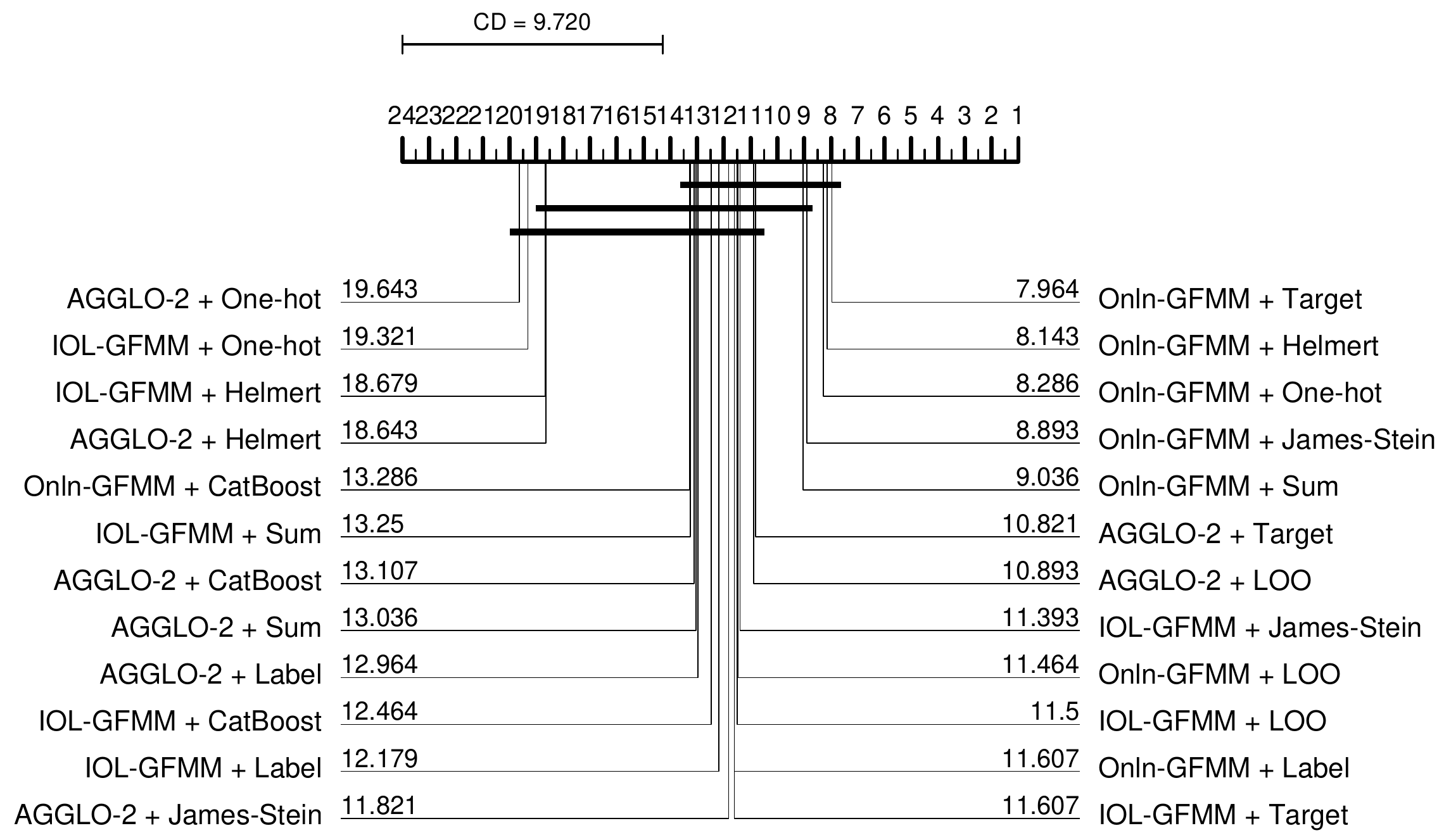}
    \caption{Critical difference diagram for the performance of encoding methods on the GFMM learning algorithms with $\theta = 0.1$}
    \label{fig_hypo_test_enc_01}
\end{figure}

For three GFMM learning algorithms using $\theta = 0.1$ and eight encoding methods, we obtain $F_F = 3.9637 > 1.5655$, so the null hypothesis is rejected. It means that there are significant differences between the average class balanced accuracy of the GFMM learning algorithms using different encoding methods. Using the Nemenyi test as a posthoc test procedure, we have a CD diagram for the performance of encoding methods on the GFMM learning algorithms with $\theta = 0.1$ as shown in Fig. \ref{fig_hypo_test_enc_01}. It is easily observed that there is a statistically significant difference in the classification performance of learning algorithms in two groups: the first group contains the Onln-GFMM algorithm using target, Helmert, and one-hot encoding methods and the second group includes AGGLO-2 and IOL-GFMM algorithms using one-hot and Helmert encoding methods. Similarly, the performance of the Onln-GFMM algorithm using the James-Stein or sum encoding method significantly outperforms those using AGGLO-2 or IOL-GFMM with the one-hot encoding method. Moreover, there are no statistically significant differences in the classification results among different encoding methods on the same learning algorithm at $\theta = 0.1$.

\begin{figure}[!ht]
    \centering
    \includegraphics[width=0.9\textwidth]{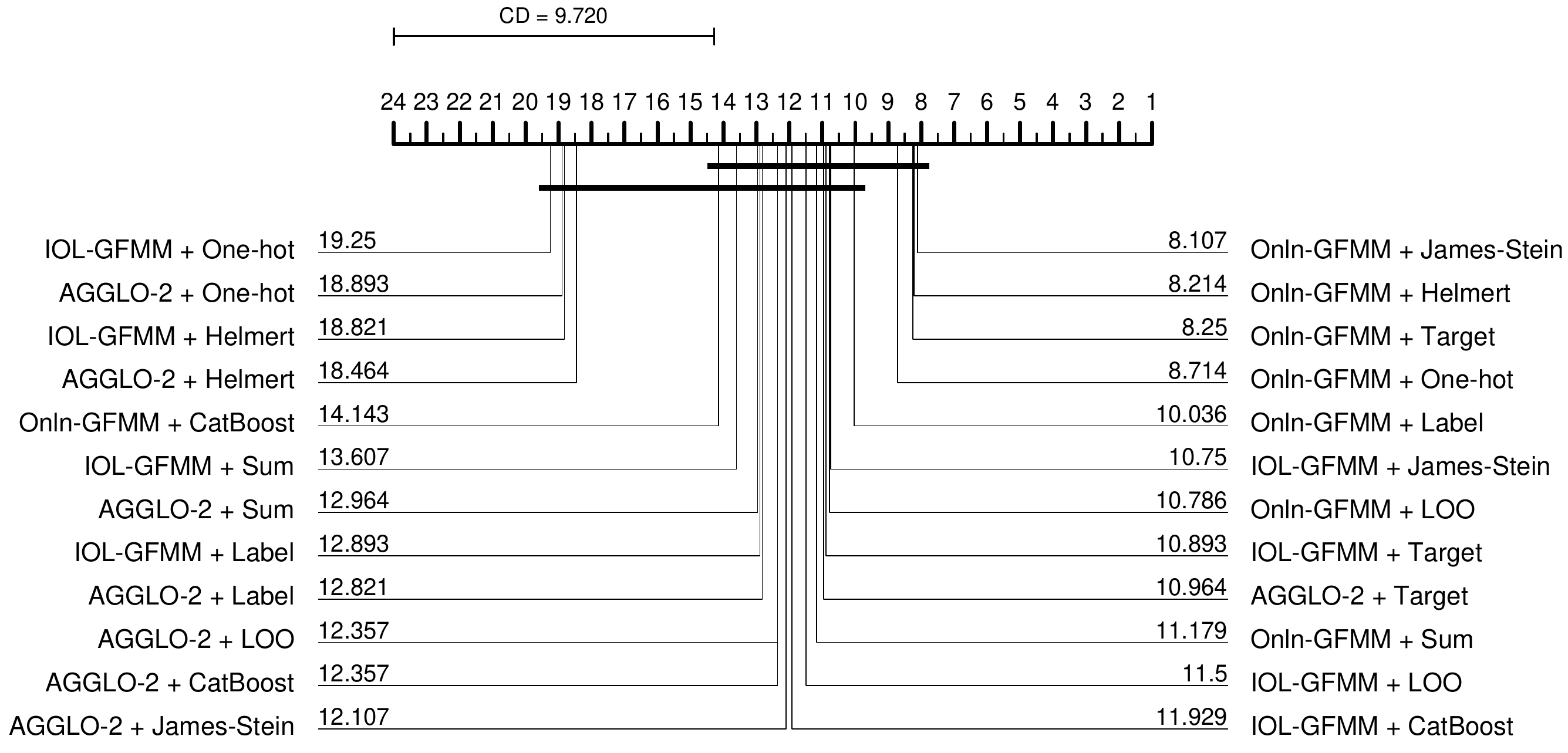}
    \caption{Critical difference diagram for the performance of encoding methods on the GFMM learning algorithms with $\theta = 0.7$}
    \label{fig_hypo_test_enc_07}
\end{figure}

For $\theta = 0.7$, we have $F_F = 3.7358 > 1.5655$, so there are significant differences in the classification performance among the GFMM models with different encoding methods. By applying the Nemenyi test, we obtain the CD diagram for different encoding methods in Fig. \ref{fig_hypo_test_enc_07}. We can see that there are statistically significant differences between the GFMM models trained by learning algorithms in two groups: the first group consists of the Onln-GFMM algorithm using James-Stein, Helmert, target, or one-hot encoding and the second group comprises IOL-GFMM or AGGLO-2 using one-hot or Helmert encoding method. Furthermore, there is no significant difference in the performance among encoding methods on the same learning algorithm at $\theta = 0.7$.

For $\theta = 1$, we obtain $F_F = 3.6582 > 1.5655$, so the null hypothesis is rejected. Using the Nemenyi post-hoc test, we receive the CD diagram in Fig. \ref{fig_hypo_test_agglo2_1}. It is easily observed that there is a statistically significant difference in terms of classification performance between the IOL-GFMM or AGGLO-2 algorithm using the sum encoding method and the AGGLO-2 algorithm using the James-Stein or target encoding technique. The classification performance of the Onln-GFMM algorithm with the James-Stein encoding method is also significantly better than that using the AGGLO-2 algorithm with the sum encoding method. In addition, there is no difference in the predictive performance among the remaining pairs of encoding methods and GFMM learning algorithms at $\theta = 1$.

\begin{figure}[!ht]
    \centering
    \includegraphics[width=0.9\textwidth]{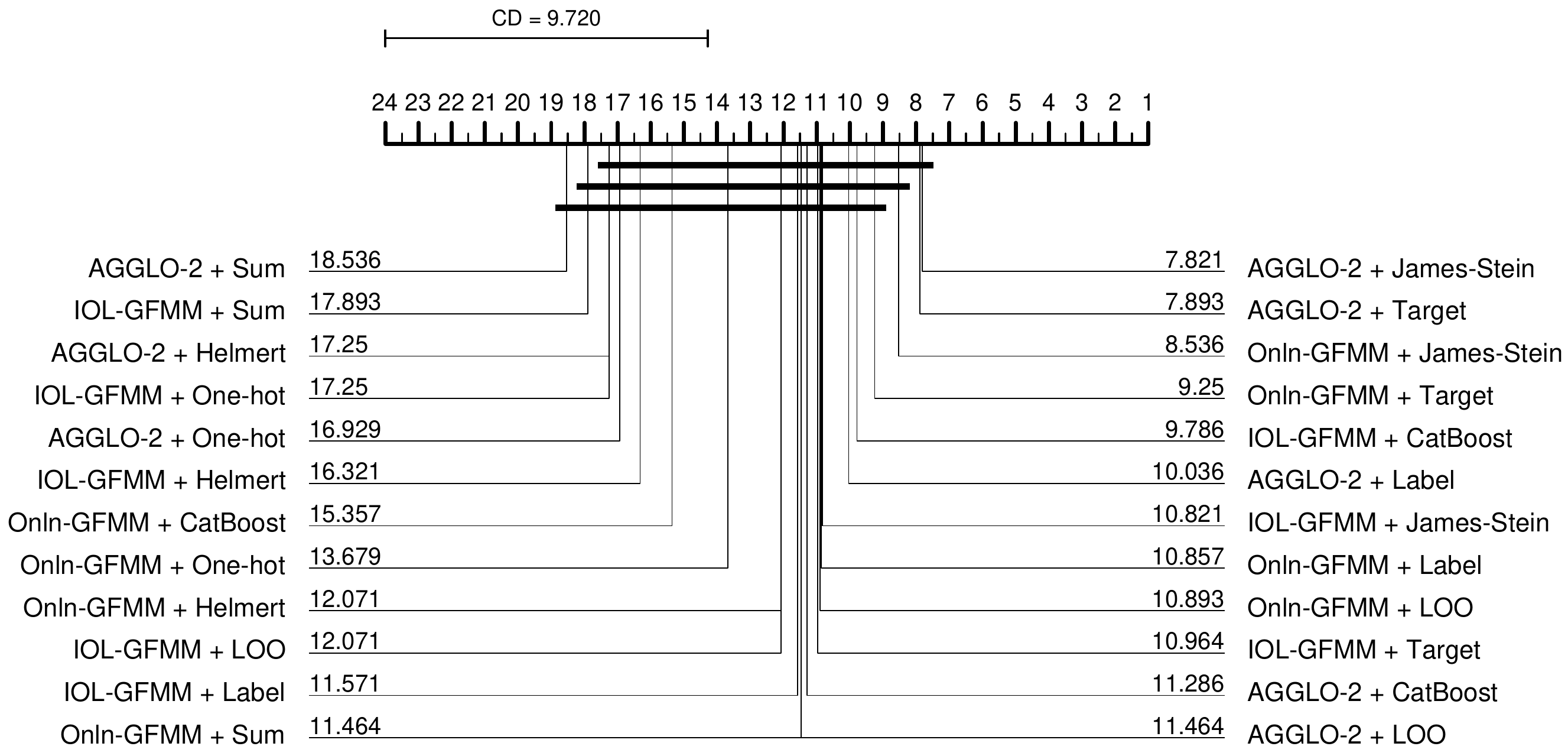}
    \caption{Critical difference diagram for the performance of encoding methods on the GFMM learning algorithms with $\theta = 1$}
    \label{fig_hypo_test_enc_1}
\end{figure}

In conclusion, the target and James-Stein encoding methods are the best encoding methods for all GFMM learning algorithms with both small and large values of the maximum hyperbox size threshold. The Helmert and one-hot encoding approaches should not be used for IOL-GFMM and AGGLO-2 algorithms. The CatBoost and LOO encoding methods with the offset in the encoded values between training samples and unseen samples are also not appropriate for all GFMM learning algorithms.

\subsection{Evaluating the effectiveness of the combination of the decision tree and the GFMM neural network}

This experiment is to assess the effectiveness of two ways of combining the GFMM model trained on numerical data different values of $\theta$ and the decision tree trained separately on categorical data. The experimental results are shown in Table \ref{table_combination_gfmm_dt} in \ref{appendix_a}. The average rank of six models at each value of $\theta$ over 11 datasets is shown in Table \ref{table_avg_rank_comb}. Only 11 mixed-type feature datasets satisfy the requirement of the experiments in this section because the combination model cannot be applied to three datasets with only categorical features. The best result in each row is highlighted in bold.

In general, it is easily observed that the classification performance of the hybrid model using both combination methods outperforms that of a single GFMM model trained on only numerical data (shown in Tables \ref{table_encoded_iol}, \ref{table_encoded_onln}, and \ref{table_encoded_agglo2}). The performance of the learning algorithms for the GFMM neural network using $\theta = 1$ is lowest, but when we combine them with the decision tree, the hybrid models can achieve relatively high performance. This fact confirms that the decision tree trained on the categorical data contributes valuable information to support the prediction ability of the hybrid model.

\begin{table}[!ht]
\centering
\scriptsize{
\caption{Average rank of the combination methods with different learning algorithms for each threshold of $\theta$} \label{table_avg_rank_comb}
\begin{tabular}{ccccccc}
\hline
\multirow{2}{*}{$\theta$} & \multicolumn{3}{c}{Using only training data} & \multicolumn{3}{c}{Using training and validation data} \\ \cline{2-7} 
                          & IOL-GFMM      & Onln-GFMM      & AGGLO-2      & IOL-GFMM          & Onln-GFMM         & AGGLO-2         \\ \hline
0.1                       &  2.7273      & 2.8636        & \textbf{2.5909}      & 4.2727           & 4.3636           & 4.1818              \\ \hline
0.7                       & \textbf{2.1818}          & 2.2727        & 3.1818      & 4.6364           & 4           & 4.7273         \\ \hline
1                         & \textbf{2.0909}      & 2.6364        & 2.5455      & 4.7273              & 3.9091              & 5.0909         \\ \hline
\end{tabular}
}
\end{table}

Of two ways of combining the decision tree and GFMM model, the classification performance of the hybrid model using only training data usually outperforms that of the model employing the training and validation data. This proves that the use of only training data to build a two-layer combination classifier still maintains an excellent classification performance and does not overfit the training data. The separation of training data into training and validation for each learning level reduces the number of training samples, and so it leads to the reduction of predictive performance. In addition, the differences in distribution between the training data and the validation data can also cause a decrease in the classification performance.

For the first approach of the combination of the GFMM model and the decision tree, the use of the IOL-GFMM algorithm gives the best classification performance among the three learning algorithms with $\theta = 0.7$ and $\theta = 1$. For $\theta = 0.1$, the best performance belongs to the use of the AGGLO-2 algorithm in combination with the decision tree. In the second approach of the combination of the GFMM neural network and decision tree, the classification performance of the combination model using the Onln-GFMM algorithm is worse than that using the IOL-GFMM or AGGLO-2 for $\theta = 0.1$. However, for larger values of $\theta$, the combination of Onln-GFMM algorithm and decision trees gives the better performance than the combination model using IOL-GFMM or AGGLO-2 for the considered datasets.

To better understand the difference in the performance of each model, we perform a statistical significance test of a hypothesis based on the average rank of the methods at each value of $\theta$. With 11 datasets and six classifiers, $F_F$ is distributed according to the F-distribution with $6 - 1 = 5$ and $(6 - 1) \cdot (11 - 1) = 50$ degrees of freedom. The critical value of $F(5, 50, 0.05)$ for the significance level $\alpha = 0.05$ is around 2.4004.

For $\theta = 0.1$, we obtain F-distribution value $F_F = 2.6222 > 2.4004$, so the null hypothesis is rejected. Using the Nemenyi test, we achieve the CD value is 2.2733. Because the difference between the
best (the combination model using AGGLO-2 with only training data) and the worst (the combination model using Onln-GFMM with both training and validation data) performing approach is smaller than that CD value, we can conclude that the post-hoc Nemenyi test
is not sufficiently powerful to detect any significant classification performance differences between different combination models using $\theta = 0.1$.

\begin{figure}[!ht]
    \centering
    \includegraphics[width=1\textwidth]{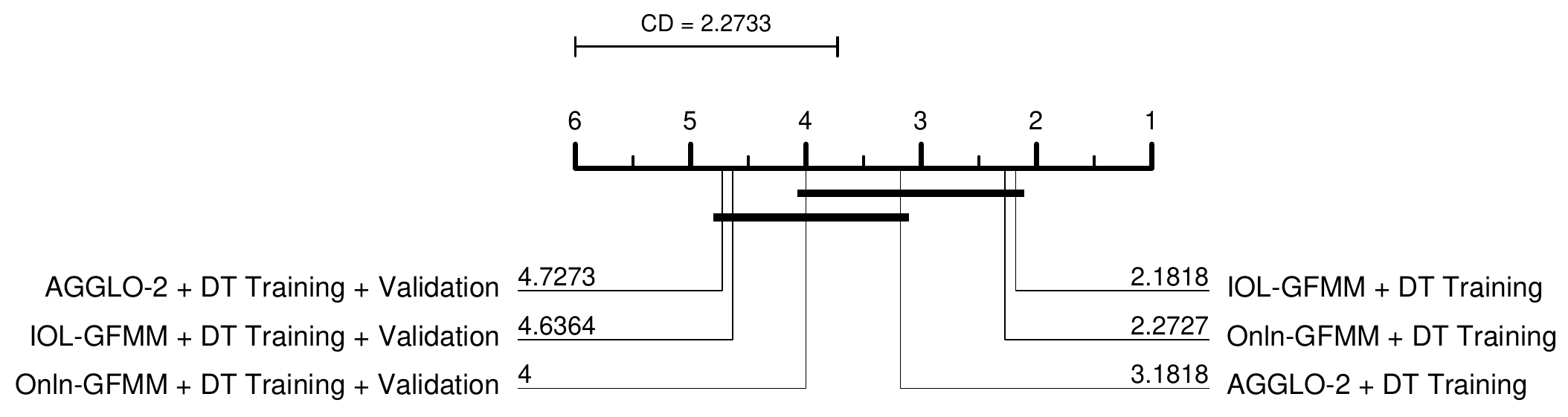}
    \caption{Critical difference diagram for the performance of combining decision tree (DT) with the GFMM models using $\theta = 0.7$}
    \label{fig_hypo_test_combine_07}
\end{figure}

For $\theta = 0.7$, we have $F_F = 5.7556 > 2.4004$, so the null hypothesis is rejected. Using the Nemenyi test, we obtain the CD diagram for six hybrid models in Fig. \ref{fig_hypo_test_combine_07}. There is a statistically significant difference in the classification performance between the combination model of the Onln-GFMM or IOL-GFMM and decision tree using only training data and hybrid models in the second combination approach using both training and validation sets with AGGLO-2 or IOL-GFMM algorithm. In addition, there are no significant differences between the remaining pairs of hybrid models.

\begin{figure}[!ht]
    \centering
    \includegraphics[width=1\textwidth]{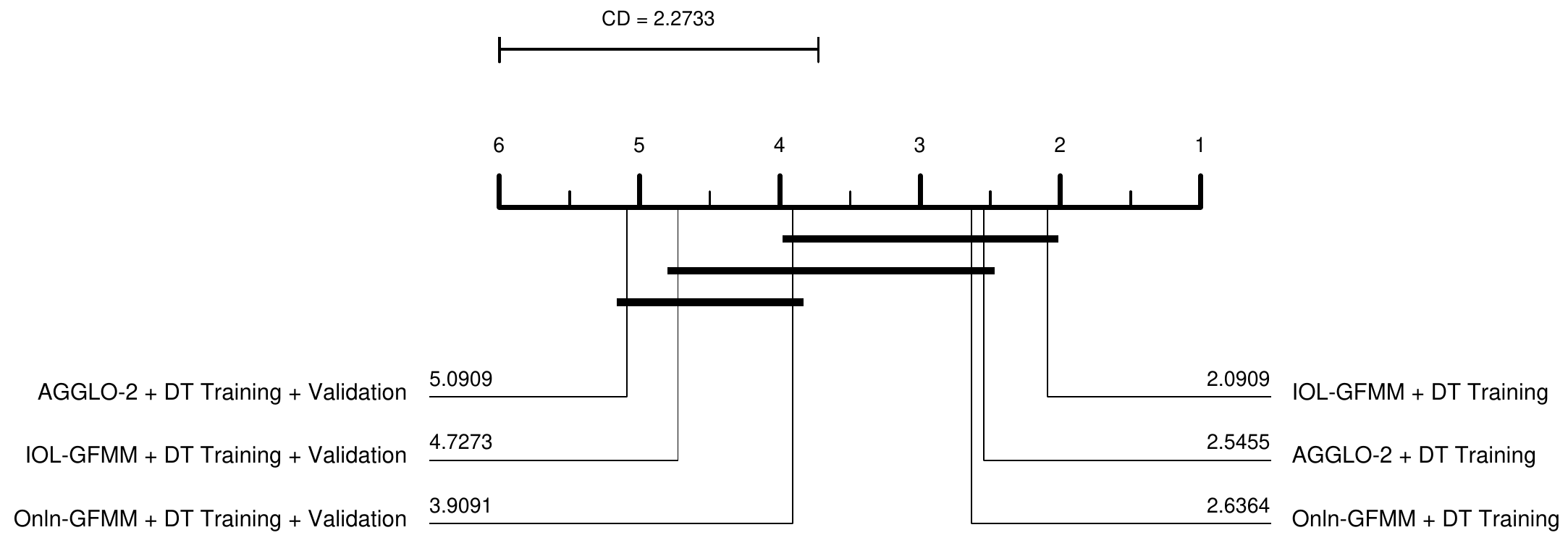}
    \caption{Critical difference diagram for the performance of combining decision tree (DT) with the GFMM models using $\theta = 1$}
    \label{fig_hypo_test_combine_1}
\end{figure}

For $\theta = 1$, we obtain the F-distribution value $F_F = 8.1304 > 2.4004$, so the null hypothesis is also rejected with a high level of significance $\alpha = 0.05$. By applying the Nemenyi test for the posthoc test, we obtain the CD diagram in Fig. \ref{fig_hypo_test_combine_1}. It is easily observed that there is a statistically significant difference in the classification accuracy between the hybrid models using only training data and the ones employing both training and validation data with AGGLO-2 algorithm to train the systems. Moreover, the performance of the hybrid model using only training data and IOL-GFMM significantly outperforms that using both training and validation data with AGGLO-2 or IOL-GFMM. However, for each combination approach, there is no significant difference between the performance of different GFMM learning algorithms.

\subsection{Evaluating the performance of algorithms with mixed features learning abilities for the GFMM neural network}
The primary purpose of this experiment is to compare the classification performance of two learning algorithms with mixed-type features learning ability for the GFMM neural network using different settings of parameters. Similarly to the above experiments, we also use three different settings for the maximum hyperbox size of numerical features $\theta = 0.1, 0.7$, and 1. For the maximum hyperbox size $\eta$ of the categorical features in the Onln-GFMM-M1 algorithm, we used the same setting as in the numerical features. For the minimum number of values of categorical features matching between the input pattern and the expandable hyperbox candidates ($\beta$) in the Onln-GFMM-M2 algorithm, we set this parameter to 25\%, 50\%, and 75\% of the number of categorical features ($r$) with $\beta \geq 1$. The average class balanced accuracy of two algorithms with different settings is shown in Table \ref{table_mix_gfmm} in \ref{appendix_a}. The best value in each row is highlighted in bold. The number of generated hyperboxes of algorithms is presented in Table \ref{table_gfmm_mix_hyperbox}.

For small values of $\eta$, it is more difficult for the hyperbox expansion process to occur, and thus there are more generated hyperboxes in the final model. For large values of $\eta$, the hyperbox expansion process depends mainly on the value of $\theta$. In the Onln-GFMM-M2 algorithm, the smaller the value of $\beta$, the easier the hyperbox expansion condition for the categorical features is met. Therefore, with a fixed value of $\theta$, the number of hyperboxes generated in the final model using a small value of $\beta$ is lower than that using a large value of $\beta$. In general, the number of generated hyperboxes in the Onln-GFMM-M1 algorithm is usually higher than that in the Onln-GFMM-M2 because the expansion conditions of the Onln-GFMM-M1 are more rigorous compared to the expansion constraints of the Onln-GFMM-M2. As a result, the complexity of the GFMM model trained by the Onln-GFMM-M1 algorithm is often higher than the model trained using the Onln-GFMM-M2.  

In terms of classification performance, with the same value of $\theta$, the Onln-GFMM-M1 is usually superior to the Onln-GFMM-M2 algorithm. These results confirm that the learning mechanism in the Onln-GFMM-M1 using the similarity degree of categorical values based on the relation of those categorical values with respect to classes is more effective than the bit matching method based on the one-hot encoding used in the Onln-GFMM-M2. However, for datasets with many features containing only two values such as \textit{flag} and \textit{zoo}, the use of the Onln-GFMM-M2 algorithm usually achieves better classification accuracy than that of the Onln-GFMM-M1. It is because the one-hot encoding mechanism and the bit matching approach for the membership function in the Onln-GFMM-M2 algorithm is more appropriate for this type of data.

\begin{table}[!ht]
\centering
\scriptsize{
\caption{The average ranking of algorithms on each value of $\theta$} \label{table_avg_rank_mix_agglo}
\begin{tabular}{ccccccc}
\hline
\multirow{2}{*}{$\theta$} & \multicolumn{3}{c}{Onln-GFMM-M1}            & \multicolumn{3}{c}{Onln-GFMM-M2}                                                \\ \cline{2-7} 
                          & $\eta = 0.1$ & $\eta   = 0.7$ & $\eta   = 1$ & $\beta   = 0.25 \cdot r $ & $\beta   = 0.5 \cdot r $ & $\beta   = 0.75 \cdot r $ \\ \hline
0.1                       & \textbf{2.5357}      & 3.6071        & 3.8571      & 4.1071                  & 3.8214                  & 3.0714                   \\ \hline
0.7                       & \textbf{1.6071}      & 2.6786        & 3.8571      & 4.2143                  & 4.5714                  & 4.0714                  \\ \hline
1                         & \textbf{1.6071}      & 2.8929        & 3.7857      & 4.0714                   & 4.2857                  & 4.3571                   \\ \hline
\end{tabular}
}
\end{table}

With the same value of $\theta$, among six GFMM models with different settings, the best predictive results are usually obtained by using the Onln-GFMM-M1 algorithm with $\eta = 0.1$. For a better comparison of the performance, we will deploy a statistical significance testing procedure as in the previous sections. Given a fixed value of $\theta$, first of all, the average ranks of models over 14 datasets are computed and shown in Table \ref{table_avg_rank_mix_agglo}. With 14 datasets and six classifiers, $F_F$ is distributed according to the F-distribution with $5$ and $65$ degrees of freedom, and so the critical value of $F(5, 65, 0.05)$ is about 2.35603.

For $\theta = 0.1$, we compute $F_F = 1.41975 < 2.35603$, so the null hypothesis is not rejected in this case. It means that for $\theta = 0.1$, the performance of both learning algorithms with considered parameter settings is not statistically different from each other.

\begin{figure}[!ht]
    \centering
    \includegraphics[width=0.75\textwidth]{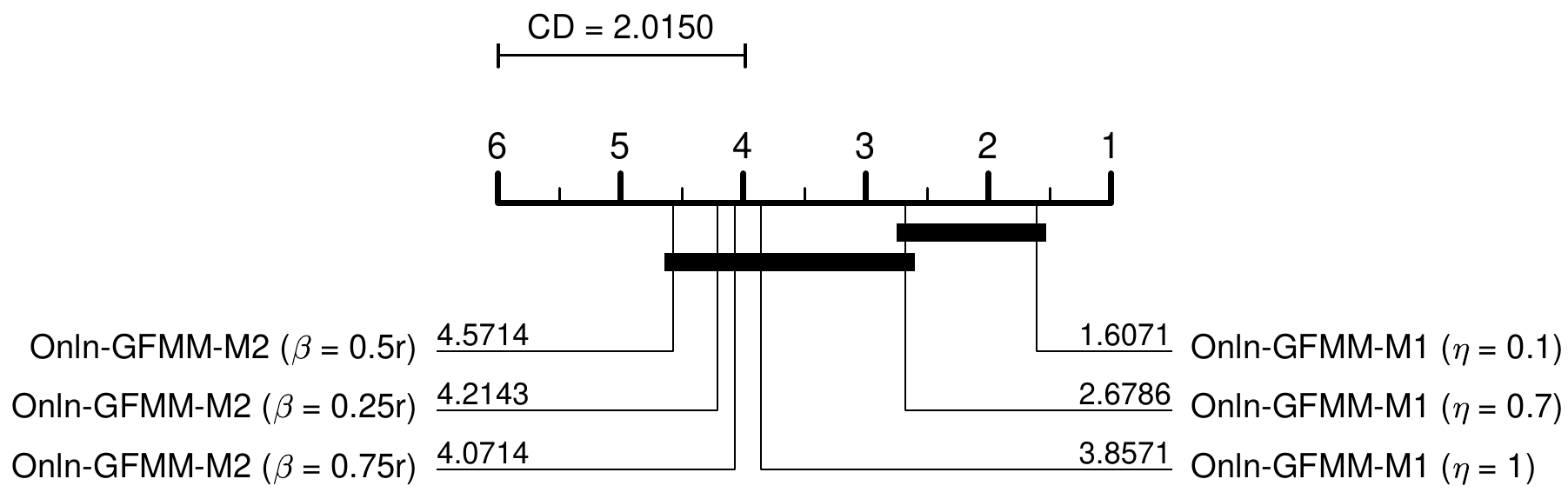}
    \caption{Critical difference diagram for the performance of learning algorithms with mixed feature learning ability using $\theta = 0.7$}
    \label{fig_mixed_gfmm_07}
\end{figure}

For a high value of $\theta = 0.7$, we have $F_F = 7.4388 > 2.35603$, so the null hypothesis is rejected. By performing the posthoc test similarly to the above experiments, we obtain the CD diagram in this case shown in Fig. \ref{fig_mixed_gfmm_07}. We can see that the Online-GFMM-M1 with $\eta = 0.1$ is statistically better than the Onln-GFMM-M1 with $\eta = 1$ and the Online-GFMM-M2 with three considered parameter settings. However, there are no statistically significant differences in the performance between the remaining pairs of models.

\begin{figure}[!ht]
    \centering
    \includegraphics[width=0.75\textwidth]{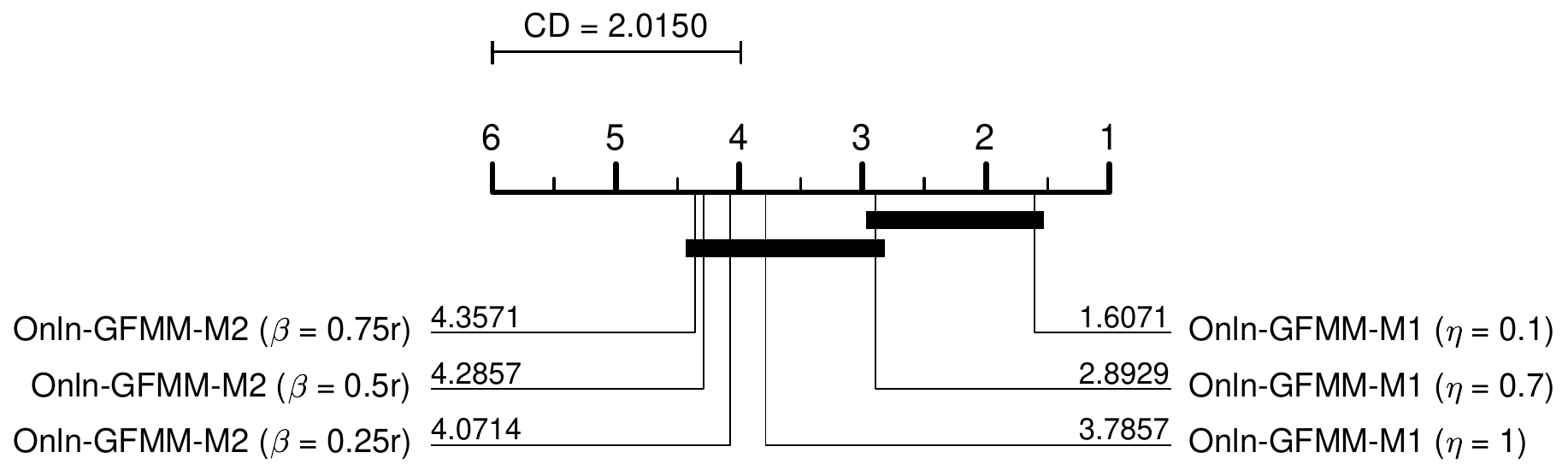}
    \caption{Critical difference diagram for the performance of learning algorithms with mixed feature learning ability using $\theta = 1$}
    \label{fig_mixed_gfmm_1}
\end{figure}

In the case of removing the expansion condition on the numerical features with $\theta = 1$, we obtain $F_F = 6.29766 > 2.35603$, and so the null hypothesis is rejected at a high level of significance $\alpha = 0.05$. In this case, we obtain a CD diagram shown in Fig. \ref{fig_mixed_gfmm_1}. We can also observe that there is only a statistically significant difference between the Onln-GFMM-M1 with $\eta = 0.1$ and Onln-GFMM-M1 ($\eta = 1$) or the Onln-GFMM-M2 with three considered parameter settings. Moreover, there are no significant differences in the classification performance among the remaining pairs of models.

\subsection{Comparison of different solutions of handling mixed-type feature datasets for the GFMM neural network}

In this experiment, we compare the best methods for handling mixed-type features data for each algorithm with a given value of $\theta$. For $\theta = 0.1$, the best encoding method for the IOL-GFMM algorithm is the James-Stein, while the best encoding method for the Onln-GFMM is the target, and the target encoding also gives the best performance for the AGGLO-2 algorithm. The use of $\eta = 0.1$ leads to the best classification results for the Onln-GFMM-M1 algorithm, while the Onln-GFMM-M2 attains the best results at $\beta = 0.75 \cdot r$. The first approach of combination using only training data results in the best results for the hybrid models. From the class balanced accuracy in each dataset, we can rank these eight methods for each dataset. The obtained results are shown in Table \ref{rank_all_01}. We used only 11 mixed-type feature datasets because three datasets with only categorical features cannot be applied to the hybrid models.

\begin{table}[!ht]
\scriptsize{
\caption{Ranking of the best methods of handling datasets with mixed-type features for the GFMM model with $\theta = 0.1$}\label{rank_all_01}
\begin{tabular}{lcccccccc}
\hline
\multirow{2}{*}{Dataset} & IOL-GFMM & Onln-GFMM & AGGLO-2 & IOL-GFMM           & Onln-GFMM           & AGGLO-2           & Onln-GFMM-M1 & Onln-GFMM-M2           \\ \cline{2-9} 
                         &  James-Stein     & Target   & Target  & \multicolumn{3}{c}{Decision tree using only   training set} & $\eta = 0.1$ & $\beta = 0.75 \cdot r$ \\ \hline
abalone          & 4     & 7              & 3     & 1 & 8     & 2     & 6     & 5     \\ \hline
australian       & 2     & 3              & 4     & 6 & 5     & 7     & 1     & 8     \\ \hline
cmc              & 3     & 4              & 2     & 7 & 8     & 6     & 5     & 1     \\ \hline
dermatology      & 7     & 1              & 6     & 2 & 3.5   & 3.5   & 5     & 8     \\ \hline
flag             & 7     & 5              & 8     & 4 & 3     & 2     & 6     & 1     \\ \hline
german           & 4     & 1              & 3     & 8 & 6     & 7     & 2     & 5     \\ \hline
heart            & 7     & 3              & 8     & 6 & 4     & 5     & 2     & 1     \\ \hline
japanese credit  & 4     & 1              & 3     & 8 & 6     & 7     & 2     & 5     \\ \hline
post operative   & 7     & 6              & 8     & 1 & 4     & 3     & 5     & 2     \\ \hline
tae              & 1     & 5              & 4     & 7 & 8     & 6     & 2     & 3     \\ \hline
zoo              & 7     & 1              & 8     & 5 & 4     & 3     & 6     & 2     \\ \hline
\textbf{Average} & 4.818 & \textbf{3.364} & 5.182 & 5 & 5.409 & 4.682 & 3.818 & 3.727 \\ \hline
\end{tabular}
}
\end{table}

We can see that, for $\theta = 0.1$, the best classification performance belongs to the Onln-GFMM algorithm using the target encoding method. Meanwhile, the combination models result in poor performance. The classification performance of the learning algorithms with mixed-type feature handling ability is also quite good in this case. For better evaluation of classification performance, we perform a hypothesis testing procedure. With 11 datasets and eight classifiers, the value of $F_F$ is distributed according to the F-distribution with $7$ and $70$ degrees of freedom. The critical value of $F(7, 70, 0.05)$ at the significance level $\alpha = 0.05$ is 2.1435. In this case, we obtain $F_F = 1.061 < 2.1435$, so the null hypothesis is not rejected. It means that at $\theta = 0.1$ there are no significant differences in the performance between eight compared classifiers.

\begin{table}[!ht]
\scriptsize{
\caption{Ranking of the best methods of handling datasets with mixed-type features for the GFMM model with $\theta = 0.7$}\label{rank_all_07}
\begin{tabular}{lcccccccc}
\hline
\multirow{2}{*}{Dataset} & IOL-GFMM & Onln-GFMM & AGGLO-2 & IOL-GFMM           & Onln-GFMM           & AGGLO-2           & Onln-GFMM-M1 & Onln-GFMM-M2             \\ \cline{2-9} 
                         & James-Stein      & James-Stein   & Target  & \multicolumn{3}{c}{Decision tree using only   training set} & $\eta = 0.1$ & $\beta = 0.75 \cdot   r$ \\ \hline
balone          & 8     & 6     & 7     & 2              & 5     & 1     & 3     & 4     \\ \hline
australian       & 3     & 7     & 2     & 4              & 1     & 6     & 5     & 8     \\ \hline
cmc              & 2     & 7     & 3     & 5              & 4     & 6     & 1     & 8     \\ \hline
dermatology      & 7     & 1     & 6     & 4              & 2.5   & 2.5   & 5     & 8     \\ \hline
flag             & 7     & 5     & 8     & 3              & 2     & 1     & 6     & 4     \\ \hline
german           & 4     & 2     & 7     & 3              & 5     & 6     & 1     & 8     \\ \hline
heart            & 3     & 1     & 5     & 6              & 4     & 7     & 2     & 8     \\ \hline
japanese credit  & 4     & 7     & 2     & 1              & 3     & 6     & 5     & 8     \\ \hline
post operative   & 7     & 2     & 6     & 1              & 3.5   & 3.5   & 5     & 8     \\ \hline
tae              & 5     & 8     & 7     & 4              & 6     & 2     & 1     & 3     \\ \hline
zoo              & 7     & 1     & 8     & 2              & 3     & 4     & 6     & 5     \\ \hline
\textbf{Average} & 5.182 & 4.273 & 5.545 & \textbf{3.182} & 3.545 & 4.091 & 3.636 & 6.545 \\ \hline
\end{tabular}
}
\end{table}

For $\theta = 0.7$, the James-Stein encoding method leads to the best performance of the IOL-GFMM and Onln-GFMM algorithms among all considered encoding techniques, while the target encoding approach helps the AGGLO-2 algorithm to obtain the best classification performance. For the Onln-GFMM-M1 algorithm, the highest performance is achieved at $\eta = 0.1$. The parameter setting $\beta = 0.75 \cdot r$ leads to the best predictive results for the Onln-GFMM-M2 algorithm. The hybrid models using the first approach of combination still attain the best performance between two combination methods. From the average class balanced accuracy of eight classifiers, we rank them over each dataset. Table \ref{rank_all_07} shows the ranking of these methods with $\theta = 0.7$.

\begin{figure}
    \centering
    \includegraphics[width=0.75\textwidth]{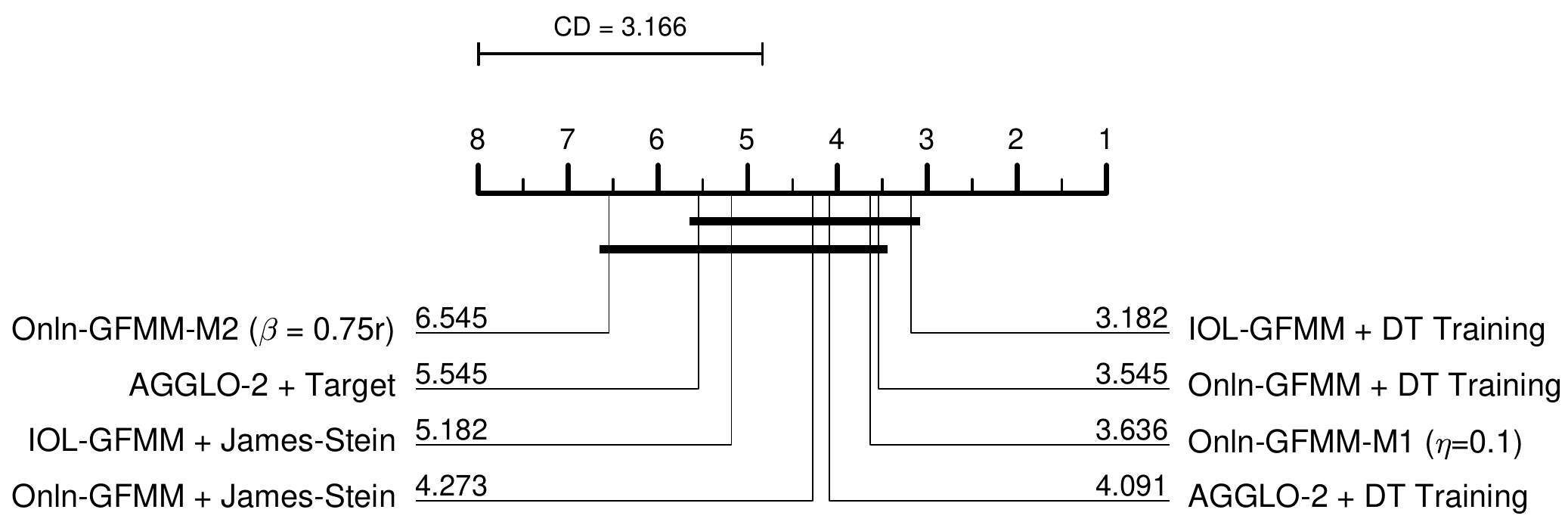}
    \caption{Critical difference diagram for the performance of the best method of handling datasets with mixed features for the GFMM model using $\theta = 0.7$}
    \label{fig_best_method_07}
\end{figure}

In contrast to the results in the case of $\theta = 0.1$, in this case, the hybrid model using the IOL-GFMM and Onln-GFMM algorithms obtains the best classification results among our proposed methods. The Onln-GFMM-M1 also leads to promising classification performance. However, the use of encoding methods for the GFMM learning algorithms and the Onln-GFMM-M2 algorithm result in poor predictive performance. Based on the average rank of methods, we conduct a statistical significance testing procedure. In this case, we have $F_F = 2.8616 > 2.1435$, so the null hypothesis is rejected. Therefore, there are statistically significant differences in the performance among methods. Using the Nemenyi test, we achieve the CD diagram for eight methods in Fig. \ref{fig_best_method_07}. It can be observed that the classification result of the hybrid model using the IOL-GFMM algorithm significantly outperforms the performance of the GFMM model trained by the Onln-GFMM-M2 algorithm. However, there are no statistically significant differences in the predictive results among the remaining pairs of classifiers. 

\begin{table}[!ht]
\scriptsize{
\caption{Ranking of the best methods of handling datasets with mixed-type features for the GFMM model with $\theta = 1$} \label{rank_all_1}
\begin{tabular}{lcccccccc}
\hline
\multirow{2}{*}{Dataset} & IOL-GFMM & Onln-GFMM & AGGLO-2 & IOL-GFMM           & Onln-GFMM           & AGGLO-2           & Onln-GFMM-M1 & Onln-GFMM-M2             \\ \cline{2-9} 
                         & CatBoost      & James-Stein   & James-Stein  & \multicolumn{3}{c}{Decision tree using only   training set} & $\eta = 0.1$ & $\beta = 0.25 \cdot   r$ \\ \hline
abalone          & 2     & 7     & 8     & 3              & 6     & 1     & 4     & 5     \\ \hline
australian       & 5     & 8     & 6     & 2              & 1     & 3     & 4     & 7     \\ \hline
cmc              & 6     & 7     & 8     & 2              & 3     & 4     & 1     & 5     \\ \hline
dermatology      & 7     & 5     & 6     & 1              & 3     & 2     & 4     & 8     \\ \hline
flag             & 8     & 5     & 7     & 3              & 2     & 1     & 6     & 4     \\ \hline
german           & 6     & 5     & 8     & 2              & 3     & 4     & 1     & 7     \\ \hline
heart            & 5     & 8     & 6     & 4              & 3     & 2     & 1     & 7     \\ \hline
japanese credit  & 5     & 8     & 6     & 1              & 2     & 4     & 3     & 7     \\ \hline
post operative   & 3     & 8     & 1     & 2              & 5     & 4     & 6     & 7     \\ \hline
tae              & 7     & 8     & 5     & 3              & 4     & 2     & 1     & 6     \\ \hline
zoo              & 6     & 1     & 7     & 4              & 2     & 3     & 8     & 5     \\ \hline
\textbf{Average} & 5.455 & 6.364 & 6.182 & \textbf{2.455} & 3.091 & 2.727 & 3.545 & 6.182 \\ \hline
\end{tabular}
}
\end{table}

In the case of using the highest value for the maximum hypebox size threshold ($\theta = 1$), if we use the encoding methods for the categorical features, then the CatBoost is the best fit for the IOL-GFMM, James-Stein is the most appropriate for the Onln-GFMM and AGGLO-2 algorithms. The Onln-GFMM-M1 achieves the best experimental outcomes at $\eta = 0.1$, while the Onln-GFMM-M2 obtained the best results with $\beta = 0.25 \cdot r$. Based on the average class balanced accuracy values in the above experiments, we can rank eight classifiers in Table \ref{rank_all_1}.

\begin{figure}
    \centering
    \includegraphics[width=0.75\textwidth]{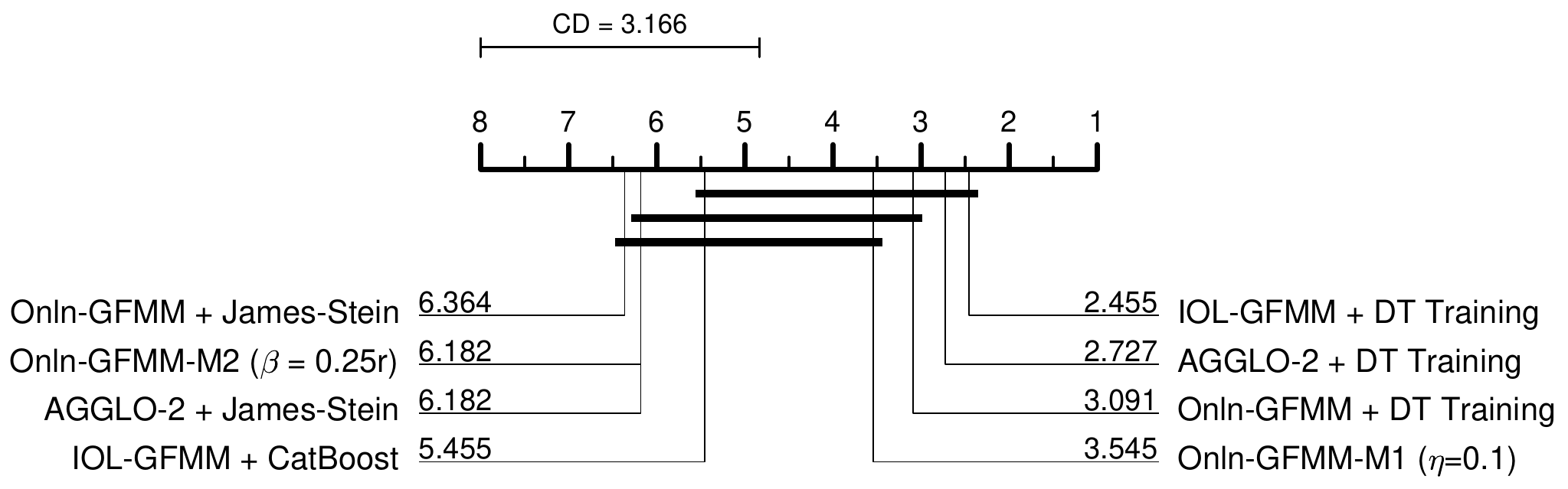}
    \caption{Critical difference diagram for the performance of the best method of handling datasets with mixed features for the GFMM model using $\theta = 1$}
    \label{fig_best_method_1}
\end{figure}

Similarly to the results for $\theta = 0.7$, the best classifiers are the hybrid models between the GFMM neural network and the decision tree. The Onln-GFMM-M1 also gives promising outcomes. Meanwhile, the poor classification performance belongs to the original learning algorithms using the encoding methods and the Onln-GFMM-M2 algorithm. To better understand the difference in the performance among classifiers, we perform a statistical testing procedure. In this case, we obtain $F_F = 9.334 > 2.1435$, so the null hypothesis is rejected. Likewise the above experiments, we achieve a CD diagram in Fig. \ref{fig_best_method_1}. In this case, we can observe that the hybrid models between decision tree and IOL-GFMM and AGGLO-2 algorithms statistically outperform the Onln-GFMM and AGGLO-2 algorithms using encoding methods as well as the Onln-GFMM-M2 algorithm. There is also a significant difference in the classification performance between the Onln-GFMM algorithm using the James-Stein encoding method and hybrid models. However, the predictive results of the hybrid models and the Onln-GFMM-M1 algorithm are similar to each other. Also, there are no significant differences in the performance among the original learning algorithm with encoding methods and the improved algorithms with mixed-type features learning ability.

\section{Existing disadvantages and potential research directions}\label{direction}
\subsection{Existing drawbacks}
One of the key strong points of the online learning algorithms for the GFMM neural network, such as the IOL-GFMM and the Onln-GFMM, is the incremental learning ability. However, the use of encoding methods (except the label and Catboost encoding techniques), the combination of the GFMM model and decision trees, and the current specific learning algorithms with mixed-type feature learning abilities need an entire training set at the first training stage. Therefore, the incremental learning ability of the GFMM model is lost, and so the GFMM model cannot be applied to streaming data with categorical features. The use of the label or CatBoost encoding method can maintain the incremental learning capability of the GFMM model, but as shown in the experiment, the performance of the GFMM model trained by the Onln-GFMM or IOL-GFMM with these two encoding methods is not high. In addition, each encoding method has its own drawbacks.

The CatBoost encoding method depends on the data presentation order to form the encoded values, so it is sensitive to the order of training samples presentation. In addition, there is a shift in the encoded value between training and testing data. With the same categorical value, its encoded value in training data is different from that in the testing data. As a result, test data may contain new encoded categorical values that do not exist in the training set. Even in the training set, a categorical value can receive many different encoded values depending on the historical data before the current training sample. This characteristic results in a poor classification performance when we use the Onln-GFMM algorithm with a hyperbox contraction process. Meanwhile, the label encoding does not face these issues, but it imposes an artificial distance between categorical values without considering any other information. The dummy encoded values for categories do not reflect the relationship between values in the same categorical feature. Therefore, the predictive results of the GFMM models using the label encoding method are not high either. Similarly to the CatBoost encoding method, the LOO encoding method also faces a shift between values in the training and testing patterns because the encoded value for a categorical value of a current sample in the training set is computed from the training data excluding that sample, while in the testing set the encoded value is computed from all training samples. Such a shift may gradually decrease the classification performance of the GFMM models, especially in the case of using the learning algorithms with the hyperbox contraction process. Another drawback of the LOO encoding technique is the overfitting issue because the encoded values are computed from class values of training samples.

The encoding methods based on simple statistical measures such as the target and James–Stein can result in overfitting and target leakage due to the use of target information to calculate the encoded values. These methods are based on the class variable, and thus they will be impacted by the noise in the training data.

The one-hot encoding, Helmert, and Sum encoding methods create many new features with many values of 0 and 1. The training data with many values of 0 and 1 in features will prevent the expansion process of hyperboxes. In addition, the current membership function of the GFMM model does not handle well the values of 0 and 1 only. Therefore, the effect of the membership function in the classification process is limited if we use these encoding methods. Another shortcoming of these types of encoding is an enormous number of added features when categorical features consist of a large number of unique categorical values. The increase in the number of features will raise the computational cost and lead to the sparse space. If we use the dimensionality reduction techniques to overcome this issue, it is at a risk of losing information \cite{Cerda18}. Another disadvantage of these encoding approaches is that they do not provide an additional heuristic solution to handle new categorical values appearing in the testing set.

The hybrid classifiers between the GFMM neural network and the decision tree have certain advantages in terms of performance and flexibility, but they also face the ambiguity in the case that training samples show different classes for the same training features. This issue results from the separation of the training data into the categorical and numerical samples. In several cases, two numerical (categorical) samples are identical to each other on all features, but they have different classes because the original samples are only different on the categorical (numerical) features. Therefore, the learned models are ambiguous when finding the correct class for an input pattern. In this case, we can obtain many hyperboxes with the same coordinates but different classes, or these hyperboxes are contracted and cause unnecessary disturbance. As a result, the classification performance deteriorates.

Two current learning algorithms for the GFMM neural network which can learn from mixed-type features also have many existing problems. Both learning algorithms need to be given the full training data at the training time to encode the categorical values or build a membership function. In the Onln-GFMM-M1, the algorithm needs to be provided with the whole training set to compute the distance measures among categorical values. If the unseen pattern contains a new categorical value which is not in the training set, the GFMM model cannot predict this sample because it does not know the distance value from this new categorical value to the existing categorical values of the current hyperboxes. Meanwhile, the Onln-GFMM-M2 needs to utilize the full training set to perform one-hot encoding for categorical features. The use of one-hot encoding for categorical features cannot handle a new categorical value using which the model has not been trained. Additionally, employing a single-pass learning through the training samples makes the constructed hyperboxes dependent on the data presentation order, and so the classification performance of these algorithms also depends on data presentation order in the same manner as in the incremental learning algorithms.

In addition sacrificing the incremental learning ability of the original learning algorithm, the Onln-GFMM-M1 and Onln-GFMM-M2 algorithms have their own problems. For the Onln-GFMM-M1 algorithm, the class conditional probabilities used to build the membership function for categorical features are unknown, and so they are estimated from the training data. This makes the obtained values likely to overfit the training set, especially when the training data are sparse or contain a small number of samples. The current distance measure between two categorical values in the same categorical feature depends only on the frequency of those values with respect to class values. The algorithm has not yet considered the correlations between categorical features. Therefore, it assumes and requires the independence between variables. The implicit assumption of feature independence sometimes leads to wrong distance value between categorical values. A simple example illustrating this issue is XOR data. The distance measure in the Onln-GFMM-M1 algorithm will generate zero distance between all categorical values, which is obviously undesirable. As a result, the classification performance will deteriorate when eliminating the significant correlations among features \cite{Cheng04}. Furthermore, in practice, variables are usually correlated with each other. For example, given a categorical feature \textit{City} containing three values \textit{Canberra}, \textit{Sydney}, \textit{London}. In the common context, value \textit{Canberra} is closer to \textit{Sydney} than \textit{London} because of the geographic distance. However, if we have another categorical feature such as \textit{Is Capital?}, then value \textit{Canberra} should be closer to \textit{London} than \textit{Sydney} in the correlation between feature \textit{City} and feature \textit{Is Capital?} as both of them are capital cities. The relationship between features can provide useful information to improve the classification performance of the classifiers. Therefore, we should compute the distance between categorical values of a categorical feature using both the correlations between that feature and classes as well as with other features. Moreover, the use of lower and upper bounds for categorical features of hyperboxes will reduce the interpretability of the GFMM model because it does not make sense when converted to relational operators. For the Onln-GFMM-M2 algorithm, it is easy to extract the rule sets for the categorical features from the one-hot encoding representation using logical operators. However, the learning algorithm and the membership function are straightforward. The membership function for the categorical features, which depends only on the number of values matching between the input pattern and the categorical features of existing hyperboxes are not strong enough to handle many cases occurring in practice. For example, given a dataset containing only categorical features, if two hyperboxes have the same number of matching values with respect to the input pattern, the membership values of these two hyperboxes on the categorical features are equal. In this case, the algorithm has to select the predicted class randomly. In addition, this method also disregards the correlation between features when building the membership function. This learning algorithm uses the one-hot encoding method for the categorical values, so it also neglects the relationship between categorical values in the same categorical attribute. As a result, the performance of the Onln-GFMM-M2 is the worst among the classifiers in the experiments.

All of the current learning algorithms only work on input samples located within the range of [0, 1]. Therefore, the classification performance of learning algorithms is influenced by the normalized values and the range of training samples. If the testing samples are outside of the range of training samples, the GFMM model is likely to yield wrong predicted results. The constraint on the range of input samples prevents the applicability of the GFMM model for streaming data in which the model is not provided with information of feature ranges in advance. 

\subsection{Potential research directions}
\textbf{Construction of the learning algorithm for streaming data}. The first research direction should focus on constructing the learning algorithms for the GFMM neural network, which can handle both categorical and numerical features in streaming data. For this purpose, new learning algorithms need the learning abilities for mixed-type features in an incremental manner. Therefore, we need to compute the distance between categorical features incrementally based on the historical data. We also have to build a new representation for categorical features of hyperboxes so that they can accommodate new categorical values in real-time.

\textbf{Building an interpretable GFMM model for both numerical and categorical features}. For both current learning algorithms of the GFMM model with the ability to learn from both categorical and numerical features, the use of lower and upper bounds for categorical features in the Onln-GFMM-M1 can achieve pretty good classification performance, but this way of representation limits the interpretability of the final model. In the Onln-GFMM-M2 algorithm, the use of one-hot encoding for each categorical feature in the hyperbox can easily generate the rules in the form of AND and OR operators to explain the predictive results on the categorical features. However, as shown in the experiments, the classification performance of the Onln-GFMM-M2 is poor. Therefore, it is necessary to build the interpretable learning algorithms for both categorical and numerical features with high prediction performance.

\textbf{Dynamic rescaling and outlier detection}. The GFMM model only works if the input samples are in the range of [0, 1] on each dimension. Hence, the input samples need to be normalized before training the GFMM model. As a result, if the testing sample is outside of the range of the training data, then the value after normalizing would fall outside the required range of [0, 1] and the classifier does not work correctly. In the online learning process on the streaming data, we also face the same problem because we may not know the range of input features in advance and input samples are usually not normalized to the range of [0, 1]. A solution is to perform a dynamic rescaling of the ranges covered by the features as shown in \cite{Salvador16}. However, this method will be affected by the outliers when the current hyperboxes are normalized to accommodate the outliers. Therefore, we need to build an effective outlier handling mechanism to decide whether the current model should be renormalized or not. Therefore, the construction of an online adaptive GFMM neural network is an exciting field of research which should receive more attention in the future. 

\section{Conclusion}\label{conclusion}
This paper presented three types of approaches for handling and learning from mixed-type features data for the GFMM neural network, i.e., using encoding methods for categorical features, combining the GFMM model trained on the numerical data and the decision tree trained on the categorical features, and applying the learning algorithms with mixed-type feature learning abilities. We have comprehensively assessed the performance of these approaches. For the encoding solution, we identified the appropriate encoding methods for each type of learning algorithm, as well as indicating the drawbacks of these methods. In general, the encoding methods are appropriate for only small values of the maximum hyperbox size parameter. For the low threshold of $\theta$, there is not much difference in the classification performance between the proposed methods. Meanwhile, the combination of the GFMM model and the decision tree is a flexible approach and able to obtain high accuracy in the case of using large values of the maximum hyperbox size. The learning algorithms with the mixed-type feature learning capabilities are potential approaches to expand the strength of the original learning algorithms for the categorical data in a natural way. Therefore, much research attention should be put on the learning algorithms for both numerical and categorical features in the future. Based on the experimental results, we have also shown the drawbacks of existing methods and proposed the potential directions for future studies.

\bibliography{mybibfile}

\appendix
\setcounter{table}{0}

\section{Details of the experimental results}\label{appendix_a}
This part provides the details of the obtained results of experiments in this paper. Tables \ref{table_encoded_iol}, \ref{table_encoded_onln}, and \ref{table_encoded_agglo2} show the average class balanced accuracy values of the IOL-GFMM, Onln-GFMM, and AGGLO-2 algorithms with different encoding methods. The complexity of the models is demonstrated by the number of generated hyperboxes. These numbers of hyperboxes are shown in Tables \ref{table_encoded_iol_hyperbox}, \ref{table_encoded_onln_hyperbox}, and \ref{table_encoded_agglo2_hyperbox}.

Table \ref{table_combination_gfmm_dt} shows the average class balance accuracy of the hybrid models combining the GFMM neural network with the decision tree using different learning algorithms. Table \ref{table_mix_gfmm} describes the experimental results of both learning algorithms for the GFMM neural network with mixed-type feature learning abilities, while Table \ref{table_gfmm_mix_hyperbox} shows the number of hyperboxes generated in the final model using these algorithms.

\begin{table}[!ht]
\scriptsize{
\caption{The average class balanced accuracy of encoding methods for the GFMM model using the IOL-GFMM algorithm} \label{table_encoded_iol}
\begin{tabular}{lcC{1.8cm}cccccccc}
\hline
Dataset                            & $\theta$ & Numerical features only & CatBoost & One-hot & LOO     & Label   & Target  & James-Stein & Helmert & Sum     \\ \hline
\multirow{3}{*}{abalone}           & 0.1      & \textbf{0.11292}                            & 0.1077                       & 0.11216                     & 0.08367                 & 0.11217                   & 0.11161                    & 0.11161                         & 0.11216                     & 0.11215                 \\ \cline{2-11} 
                                   & 0.7      & 0.10569                                     & \textbf{0.11117}             & 0.04521                     & 0.04599                 & 0.07214                   & 0.0544                     & 0.0544                          & 0.04521                     & 0.03942                 \\ \cline{2-11} 
                                   & 1        & 0.10565                                     & \textbf{0.11041}             & 0.03651                     & 0.05247                 & 0.07843                   & 0.07027                    & 0.07027                         & 0.02491                     & 0.02491                 \\ \hline
\multirow{3}{*}{australian}        & 0.1      & 0.64903                                     & 0.79148                      & 0.6283                      & 0.70876                 & 0.77303                   & 0.77283                    & 0.77321                         & 0.62855                     & \textbf{0.79163}        \\ \cline{2-11} 
                                   & 0.7      & 0.68137                                     & \textbf{0.81259}             & 0.63919                     & 0.60169                 & 0.79844                   & 0.80302                    & 0.80712                         & 0.63943                     & 0.80088                 \\
\hline
\end{tabular}
}
\end{table}

\begin{table}[!ht]
\scriptsize{
\caption{The number of hyperboxes generated by the GFMM model using the IOL-GFMM algorithm with different encoding methods} \label{table_encoded_iol_hyperbox}
\begin{tabular}{L{2.7cm}cC{1.8cm}cccccccc}
\hline
\shortstack[l]{Dataset\\(No. training samples)}            & $\theta$ & Numerical features only & CatBoost & One-hot & LOO     & Label   & Target   & James-Stein & Helmert & Sum     \\ \hline
\multirow{3}{*}{abalone (3313)}           & 0.1      & 1418.6                                      & 1387.15                      & 704.55                      & 714.875                 & 704.7                     & 709.75                     & 709.75                          & 704.55                      & 704.65                  \\ \cline{2-11} 
                                          & 0.7      & 1523.175                                    & 1432.4                       & 79.975                      & 343.575                 & 490.25                    & 244.725                    & 244.725                         & 79.975                      & 57.725                  \\ \cline{2-11} 
                                          & 1        & 1522.525                                    & 1434.7                       & 43.75                       & 412.525                 & 502.95                    & 679.725                    & 679.725                         & 26.75                       & 26.75                   \\ \hline
\multirow{3}{*}{australian (518)}         & 0.1      & 212.675                                     & 460.575                      & 493.875                     & 468.075                 & 478                       & 472.7                      & 467.95                          & 493.875                     & 493.875                 \\ \cline{2-11} 
                                          & 0.7      & 44.975                                      & 44.7                         & 330.475                     & 55.425                  & 70.475                    & 78.825                     & 63.35                           & 330.475                     & 55                      \\
\hline
\end{tabular}
}
\end{table}

\begin{table}[!ht]
\scriptsize{
\caption{The average class balanced accuracy of encoding methods for the GFMM model using the Onln-GFMM algorithm} \label{table_encoded_onln}
\begin{tabular}{lcC{1.8cm}cccccccc}
\hline
Dataset                            & $\theta$ & Numerical features only & CatBoost & One-hot & LOO     & Label   & Target  & James-Stein & Helmert & Sum     \\ \hline
\multirow{3}{*}{abalone}           & 0.1                   & 0.10004          & 0.09965          & 0.10033          & 0.0814           & \textbf{0.10036} & 0.09895          & 0.09895          & 0.10033          & 0.10028          \\ \cline{2-11} 
                                   & 0.7                   & \textbf{0.08991} & 0.08581          & 0.06639          & 0.05247          & 0.06577          & 0.07034          & 0.07034          & 0.06639          & 0.0625           \\ \cline{2-11} 
                                   & 1                     & \textbf{0.07758} & 0.07582          & 0.05105          & 0.07688          & 0.07701          & 0.07482          & 0.07482          & 0.04491          & 0.04605          \\ \hline
\multirow{3}{*}{australian}        & 0.1                   & 0.6327           & 0.78676          & 0.78763          & 0.72458          & 0.77516          & 0.77245          & 0.76915          & 0.78221          & \textbf{0.78788} \\ \cline{2-11} 
                                   & 0.7                   & 0.55458          & 0.7533           & \textbf{0.79609} & 0.64756          & 0.75797          & 0.75908          & 0.75103          & 0.79324          & 0.7419           \\ \cline{2-11} 
                                   & 1                     & 0.55443          & 0.55453          & 0.36412          & 0.44845          & \textbf{0.55575} & 0.55408          & 0.55501          & 0.36138          & 0.3364           \\ \hline
\multirow{3}{*}{cmc}               & 0.1                   & 0.34733          & 0.27229          & \textbf{0.41127} & 0.39216          & 0.39799          & 0.40479          & 0.40407          & 0.40834          & 0.40817          \\
\hline
\end{tabular}
}
\end{table}

\begin{table}[!ht]
\scriptsize{
\caption{The number of hyperboxes generated by the GFMM model using the Onln-GFMM algorithm with different encoding methods} \label{table_encoded_onln_hyperbox}
\begin{tabular}{L{2.7cm}cC{1.8cm}cccccccc}
\hline
\shortstack[l]{Dataset\\(No. training samples)}            & $\theta$ & Numerical features only & CatBoost & One-hot & LOO     & Label   & Target   & James-Stein & Helmert & Sum     \\ \hline
\multirow{3}{*}{abalone (3133)}           & 0.1      & 490.6     & 739.875  & 708.25  & 692.4    & 708.175 & 695.25   & 695.25      & 708.25  & 708.15  \\ \cline{2-11} 
                                          & 0.7      & 37.875    & 48.775   & 79.45   & 56.6     & 56.55   & 56.6     & 56.6        & 79.45   & 57.55   \\ \cline{2-11} 
                                          & 1        & 26.75     & 26.75    & 26.75   & 26.75    & 26.75   & 26.75    & 26.75       & 26.75   & 26.75   \\ \hline
\multirow{3}{*}{australian (518)}         & 0.1      & 196.775   & 459.9    & 493.225 & 467.625  & 478.225 & 473.325  & 467.75      & 493.225 & 493.225 \\ \cline{2-11} 
                                          & 0.7      & 11.525    & 29.725   & 330.6   & 57.05    & 70.925  & 81.3     & 66.275      & 330.6   & 54.925  \\ \cline{2-11} 
                                          & 1        & 2         & 2        & 2       & 2        & 2       & 2        & 2           & 2       & 2       \\ \hline
\multirow{3}{*}{cmc (1105)}               & 0.1      & 115.4     & 626.55   & 881.525 & 784.925  & 881.525 & 797.425  & 794.95      & 881.525 & 881.575 \\
\hline
\end{tabular}
}
\end{table}

\begin{table}[!ht]
\scriptsize{
\caption{The average class balanced accuracy of encoding methods for the GFMM model using the AGGLO-2 algorithm} \label{table_encoded_agglo2}
\begin{tabular}{lcC{1.8cm}cccccccc}
\hline
Dataset                            & $\theta$ & Numerical features only & CatBoost & One-hot & LOO     & Label   & Target  & James-Stein & Helmert & Sum     \\ \hline
\multirow{3}{*}{abalone} & 0.1 & 0.11353 & 0.10621 & \textbf{0.11395} & 0.08769 & \textbf{0.11395} & 0.11246 & 0.11246 & \textbf{0.11395} & 0.11354 \\ \cline{2-11} 
 & 0.7 & \textbf{0.11548} & 0.11043 & 0.0521 & 0.04415 & 0.05329 & 0.05522 & 0.05522 & 0.0521 & 0.0512 \\ \cline{2-11} 
 & 1 & \textbf{0.11526} & 0.11028 & 0.03798 & 0.0447 & 0.04602 & 0.05543 & 0.05543 & 0.02383 & 0.02383 \\ \hline
\multirow{3}{*}{australian} & 0.1 & 0.6472 & 0.78525 & 0.63278 & 0.72432 & 0.77621 & 0.77069 & 0.76875 & 0.63302 & \textbf{0.79496} \\ \cline{2-11} 
 & 0.7 & 0.68346 & 0.8065 & 0.64606 & 0.6607 & 0.80441 & \textbf{0.80778} & 0.79873 & 0.6463 & 0.80197 \\ \cline{2-11} 
 & 1 & 0.69223 & 0.70539 & 0.36395 & 0.47553 & 0.69984 & \textbf{0.70699} & 0.69304 & 0.36381 & 0.36401 \\ \hline
\multirow{3}{*}{cmc} & 0.1 & 0.41026 & 0.27321 & 0.39802 & 0.40026 & 0.40733 & \textbf{0.41583} & 0.41479 & 0.39802 & 0.40956 \\
\hline
\end{tabular}
}
\end{table}

\begin{table}[!ht]
\scriptsize{
\caption{The number of hyperboxes generated by the GFMM model using the AGGLO-2 algorithm with different encoding methods} \label{table_encoded_agglo2_hyperbox}
\begin{tabular}{L{2.7cm}cC{1.8cm}cccccccc}
\hline
\shortstack[l]{Dataset\\(No. training samples)}            & $\theta$ & Numerical features only & CatBoost & One-hot & LOO     & Label   & Target   & James-Stein & Helmert & Sum     \\ \hline
\multirow{3}{*}{abalone (3133)}           & 0.1      & 1262.35   & 1347.525 & 741.45  & 740.4    & 741.125  & 740.975  & 740.975     & 741.45  & 740.95  \\ \cline{2-11} 
                                          & 0.7      & 1197.525  & 1215.775 & 78.75   & 192.625  & 84       & 166.175  & 166.175     & 78.75   & 56.375  \\ \cline{2-11} 
                                          & 1        & 1197      & 1214.95  & 43.275  & 195.425  & 73.075   & 164.425  & 164.425     & 26.75   & 26.75   \\ \hline
\multirow{3}{*}{australian (518)}         & 0.1      & 220.05    & 460.3    & 493.2   & 467.9    & 478.2    & 473.575  & 467.9       & 493.2   & 493.2   \\ \cline{2-11} 
                                          & 0.7      & 82.175    & 55.325   & 330.6   & 58.3     & 71.8     & 82.15    & 67.175      & 330.6   & 54.775  \\ \cline{2-11} 
                                          & 1        & 78.375    & 36.3     & 2.95    & 4.975    & 9.4      & 12.85    & 11.15       & 2       & 2       \\ \hline
\multirow{3}{*}{cmc (1105)}               & 0.1      & 171.375   & 640.825  & 880.475 & 786.225  & 880.7    & 798.75   & 795.875     & 880.475 & 880.65  \\
\hline
\end{tabular}
}
\end{table}

\begin{table}[!ht]
\scriptsize{
\caption{Experimental results of GFMM learning algorithms using different encoding methods for synthetic datasets ($\theta = 0.1$)} \label{table_syn_theta01}
\begin{tabular}{L{1.5cm}cL{3cm}cccccccc}
\hline
Dataset                       & Algorithm                  & Measure                  & CatBoost & One-hot & LOO   & Label & Target & Jame-Stein & Helmert & Sum   \\ \hline
\multirow{12}{*}{synthetic-1} & \multirow{4}{*}{IOL-GFMM}  & Accuracy                 & 0.87     & 0.854   & 0.805 & 0.854 & 0.854  & 0.854      & 0.854   & 0.854 \\ \cline{3-11} 
                              &                            & No boxes                 & 100      & 74      & 74    & 74    & 74     & 74         & 74      & 74    \\ \cline{3-11} 
                              &                            & No 2 criterion         & 3        & 6       & 0     & 6     & 6      & 6          & 6       & 6     \\ \cline{3-11} 
                              &                            & No 2 criterion correct & 3        & 3       & 0     & 3     & 3      & 3          & 3       & 3     \\ \cline{2-11} 
                              & \multirow{4}{*}{Onln-GFMM} & Accuracy                 & 0.86     & 0.855   & 0.805 & 0.855 & 0.855  & 0.855      & 0.855   & 0.855 \\ \cline{3-11} 
                              &                            & No boxes                 & 91       & 74      & 74    & 74    & 74     & 74         & 74      & 74    \\ \cline{3-11} 
                              &                            & No 2 criterion         & 4        & 6       & 0     & 6     & 6      & 6          & 6       & 6     \\
\hline
\end{tabular}
}
\end{table}

\begin{table}
\scriptsize{
\caption{Experimental results of GFMM learning algorithms using different encoding methods for synthetic datasets ($\theta = 0.7$)} \label{table_syn_theta07}
\begin{tabular}{L{1.5cm}cL{3cm}cccccccc}
\hline
Dataset                       & Algorithm                  & Measure                  & CatBoost & One-hot & LOO   & Label & Target & Jame-Stein & Helmert & Sum   \\ \hline
\multirow{12}{*}{synthetic-1} & \multirow{4}{*}{IOL-GFMM}  & Accuracy                 & 0.899    & 0.808   & 0.726 & 0.808 & 0.808  & 0.808      & 0.808   & 0.808 \\ \cline{3-11} 
                              &                            & No boxes                 & 38       & 6       & 6     & 6     & 6      & 6          & 6       & 6     \\ \cline{3-11} 
                              &                            & No 2 criterion         & 0        & 366     & 0     & 366   & 366    & 366        & 366     & 366   \\ \cline{3-11} 
                              &                            & No 2 criterion correct & 0        & 198     & 0     & 198   & 198    & 198        & 198     & 198   \\ \cline{2-11} 
                              & \multirow{4}{*}{Onln-GFMM} & Accuracy                 & 0.799    & 0.868   & 0.726 & 0.868 & 0.868  & 0.868      & 0.868   & 0.868 \\ \cline{3-11} 
                              &                            & No boxes                 & 6        & 6       & 6     & 6     & 6      & 6          & 6       & 6     \\ \cline{3-11} 
                              &                            & No 2 criterion         & 0        & 366     & 0     & 366   & 366    & 366        & 366     & 366   \\
\hline
\end{tabular}
}
\end{table}

\begin{table}[!ht]
\scriptsize{
\caption{Experimental results of GFMM learning algorithms using different encoding methods for synthetic datasets ($\theta = 1$)} \label{table_syn_theta1}
\begin{tabular}{L{1.5cm}cL{3cm}cccccccc}
\hline
Dataset                       & Algorithm                  & Measure                  & CatBoost & One-hot & LOO   & Label & Target & Jame-Stein & Helmert & Sum   \\ \hline
\multirow{12}{*}{synthetic-1} & \multirow{4}{*}{IOL-GFMM}  & Accuracy                 & 0.902    & 0.829   & 0.861 & 0.879 & 0.886  & 0.886      & 0.879   & 0.886 \\ \cline{3-11} 
                              &                            & No boxes                 & 38       & 3       & 8     & 21    & 24     & 24         & 21      & 24    \\ \cline{3-11} 
                              &                            & No 2 criterion         & 0        & 485     & 0     & 203   & 284    & 284        & 203     & 284   \\ \cline{3-11} 
                              &                            & No 2 criterion correct & 0        & 330     & 0     & 145   & 223    & 223        & 145     & 223   \\ \cline{2-11} 
                              & \multirow{4}{*}{Onln-GFMM} & Accuracy                 & 0.892    & 0.892   & 0.873 & 0.892 & 0.892  & 0.892      & 0.892   & 0.892 \\ \cline{3-11} 
                              &                            & No boxes                 & 2        & 2       & 2     & 2     & 2      & 2          & 2       & 2     \\ \cline{3-11} 
                              &                            & No 2 criterion         & 0        & 0       & 0     & 0     & 0      & 0          & 0       & 0     \\
\hline
\end{tabular}
}
\end{table}

\begin{table}[!ht]
\centering
\scriptsize{
\caption{The average class balanced accuracy of two approaches for combining the GFMM neural network and the decision tree using different original learning algorithms}\label{table_combination_gfmm_dt}
\begin{tabular}{lccccccc}
\hline
\multirow{2}{*}{Dataset}           & \multirow{2}{*}{$\theta$} & \multicolumn{3}{c}{Using only training data} & \multicolumn{3}{c}{Using training and validation data} \\ \cline{3-8} 
                                   &                           & IOL-GFMM      & Onln-GFMM      & AGGLO-2      & IOL-GFMM          & Onln-GFMM         & AGGLO-2         \\ \hline
\multirow{3}{*}{abalone}           & 0.1                       & \textbf{0.11361}         & 0.09824          & 0.113   & 0.08055           & 0.07177          & 0.07246          \\ \cline{2-8} 
                                   & 0.7                       & 0.10184          & 0.08582          & \textbf{0.11122} & 0.08134           & 0.0859           & 0.07317          \\ \cline{2-8} 
                                   & 1                         & 0.10181          & 0.07964          & \textbf{0.11101} & 0.08139           & 0.08496          & 0.07326          \\ \hline
\multirow{3}{*}{australian}        & 0.1                       & 0.64799          & 0.65996          & 0.6472           & 0.78075  & \textbf{0.79729} & \textbf{0.79729} \\ \cline{2-8} 
                                   & 0.7                       & 0.805          & \textbf{0.81329} & 0.79009          & 0.77286           & 0.7992           & 0.79932          \\ \cline{2-8} 
                                   & 1                         & 0.80514          & \textbf{0.81278} & 0.79746          & 0.77889           & 0.80049          & 0.79685          \\
\hline
\end{tabular}
}
\end{table}

\begin{table}[!ht]
\centering
\scriptsize{
\caption{The average class balanced accuracy of two types of learning algorithms for the GFMM model with mixed features} \label{table_mix_gfmm}
\begin{tabular}{lccccccc}
\hline
\multirow{2}{*}{Dataset}          & \multirow{2}{*}{$\theta$} & \multicolumn{3}{c}{Onln-GFMM-M1}                                                                 & \multicolumn{3}{c}{Onln-GFMM-M2}                                                                                           \\ \cline{3-8} 
                                  &                            & $\eta = 0.1$                            & $\eta   = 0.7$ & $\eta   = 1$                           & $\beta   = 0.25 \cdot r $               & $\beta   = 0.5 \cdot r $                & $\beta   = 0.75 \cdot r $               \\ \hline
\multirow{3}{*}{abalone}                                  & 0.1                        & 0.10182                                 & 0.10156        & \textbf{0.1039} & 0.10195                                 & 0.10195                                 & 0.10195                                 \\ \cline{2-8} 
                                  & 0.7                        & \textbf{0.09182}                        & 0.08919        & 0.07923                                & 0.08662                                 & 0.08662                                 & 0.08662                                 \\ \cline{2-8} 
         & 1                          & \textbf{0.08831}                        & 0.08357        & 0.0729                                 & 0.08455                                 & 0.08455                                 & 0.08455                                 \\ \hline
\multirow{3}{*}{australian}       & 0.1                        & \textbf{0.77345}                        & 0.76536        & 0.75532                                & 0.62766                                 & 0.6276                                  & 0.6425                                  \\
\hline
\end{tabular}
}
\end{table}

\begin{table}[!ht]
\centering
\scriptsize{
\caption{The number of hyperboxes generated by learning algorithms with mixed feature learning abilities} \label{table_gfmm_mix_hyperbox}
\begin{tabular}{lccccccc}
\hline
\multirow{2}{*}{Dataset}            & \multirow{2}{*}{$\theta$} & \multicolumn{3}{c}{Onln-GFMM-M1}            & \multicolumn{3}{c}{Onln-GFMM-M2}                                                \\ \cline{3-8} 
                                    &                           & $\eta = 0.1$ & $\eta   = 0.7$ & $\eta   = 1$ & $\beta   = 0.25 \cdot r $ & $\beta   = 0.5 \cdot r $ & $\beta   = 0.75 \cdot r $ \\ \hline
\multirow{3}{*}{abalone}            & 0.1                       & 693.95       & 591.725        & 522.825      & 704.15                    & 704.15                   & 704.15                    \\ \cline{2-8} 
                                    & 0.7                       & 77.225       & 56.875         & 46.75        & 79.375                    & 79.375                   & 79.375                    \\ \cline{2-8} 
                                    & 1                         & 63.3         & 45.275         & 37.375       & 65.275                    & 65.275                   & 65.275                    \\ \hline
\multirow{3}{*}{australian}         & 0.1                       & 472.15       & 348.225        & 222.6        & 198.775                   & 206.05                   & 279.025                   \\ \cline{2-8} 
                                    & 0.7                       & 260.275      & 95.825         & 42.7         & 11.675                    & 12.875                   & 21.875                    \\ \cline{2-8} 
                                    & 1                         & 253.25       & 86.4           & 35.725       & 2                         & 2.675                    & 6.95                      \\
\hline
\end{tabular}
}
\end{table}

\end{document}